\documentclass[acmsmall]{acmart}

\renewcommand\footnotetextcopyrightpermission[1]{}
\usepackage[utf8]{inputenc}
\usepackage{booktabs} 

\usepackage[ruled]{algorithm2e} 
\usepackage{booktabs}
\usepackage{xcolor}
\usepackage{color}
\usepackage{soul}
\usepackage{tabu}
\usepackage{courier}
\usepackage{url}
\usepackage{graphicx}
\usepackage{color}
\usepackage{array}
\usepackage{subfigure}
\usepackage[T1]{fontenc}
\usepackage{aecompl}
\usepackage{multirow}
\usepackage{url}
\usepackage{booktabs}
\usepackage{threeparttable}
\usepackage{babel}

\usepackage{bm}
\def\ie{\textit{i.e.}\xspace}

\def\etc{\textit{etc.}\xspace}

\SetAlFnt{\small}
\SetAlCapFnt{\small}
\SetAlCapNameFnt{\small}
\SetAlCapHSkip{0pt}
\IncMargin{-\parindent}

\acmJournal{TIST}
\acmVolume{0}
\acmNumber{0}
\acmArticle{0}
\acmYear{2019}
\acmMonth{8}
\acmArticleSeq{0}
\usepackage{makecell}
\usepackage{amsmath}
\usepackage{hhline} 
\usepackage{multirow} 
\usepackage{amsmath} 
\usepackage{xcolor}

\begin{document}

\title {Causal Discovery and Inference towards Urban Elements and Associated Factors}

\author{Tao Feng}
\authornotemark[1]
\email{ft19@tsinghua.org.cn}
\author{Yunke Zhang}
\authornote{Feng Tao and Yunke Zhang have equal contribution to this research and share the first-authorship.}
\email{zyk21@mails.tsinghua.edu.cn}
\affiliation{%
  \institution{Department of Electronic Engineering, Tsinghua University}
  \city{Beijing}
  \country{China}
}

\author{Xiaochen Fan}
\authornotemark[2]
    \affiliation{%
      \institution{Institute for Electronics and Information Technology in Tianjin, Tsinghua University}
      \city{Tianjin}
      \country{China}}
   \affiliation{%
     \institution{Department of Electronic Engineering, Tsinghua University}
     \city{Beijing}
     \country{China}}
\email{fanxiaochen33@gmail.com}

\author{Huandong Wang}
\email{wanghuandong@tsinghua.edu.cn}
   \affiliation{%
     \institution{Department of Electronic Engineering, Tsinghua University}
     \city{Beijing}
     \country{China}}

\author{Yong Li}
\authornote{Yong Li is the co-corresponding author.}
\email{liyong07@tsinghua.edu.cn}
\affiliation{%
  \institution{Department of Electronic Engineering, Tsinghua University}
  \city{Beijing}
  \country{China}
}

\begin{abstract}
Urban space is composed of complex interactions. \textit{Citizens}, as the city’s most essential element, deeply interacts with urban \textit{Locations} and produce \textit{Mobility} behaviors.
To uncover the city's fundamental functioning mechanisms,
it is important to acquire a deep understanding of complicated relationships
among citizens, location, and mobility behaviors.
Previous research studies have applied direct correlation analysis to investigate such relationships. Nevertheless, due to the ubiquitous confounding effects, empirical correlation analysis may not accurately reflect underlying causal relationships among basic urban elements.
In this paper, we propose a novel urban causal computing framework to comprehensively explore causalities and confounding effects among a variety of factors across different types of urban elements. In particular, we design a reinforcement learning algorithm to discover the potential causal graph, which depicts the causal relations between urban factors. The causal graph further serves as the guidance for estimating causal effects between pair-wise urban factors by propensity score matching. After removing the confounding effects from correlations, we leverage significance levels of causal effects in downstream urban mobility prediction tasks. Experimental studies on open-source urban datasets show that the discovered causal graph demonstrates a hierarchical structure, where citizens affect locations, and they both cause changes in urban mobility behaviors. Experimental results in urban mobility prediction tasks further show that the proposed method can effectively reduce confounding effects and enhance performance of urban computing tasks.
\end{abstract}

\ccsdesc[500]{Human-centered computing~Empirical studies in collaborative and social computing}
\ccsdesc[500]{Applied computing~Transportation}
\ccsdesc[500]{Information systems~Association rules}

\keywords{Urban Computing, Causal Discovery, Mobile Computing}
\maketitle


\renewcommand{\shortauthors}{Feng}

\section{Introduction} \label{sec:intro}
According to the United Nation's report, by 2050, urban population will increase to over 6.7 billion, accounting for 68\% of the world’s population (\textit{i.e.}, more than twice of rural population)~\cite{un2018world}.
Accordingly, urban space has become the most essential scenario in the field of intelligent systems and technology~\cite{bettencourt2010unified}.
In particular, urban scientists consider cities as self-organizing systems with thousands of subtly inter-connected factors~\cite{portugali2000self,bettencourt2013origins},
with dynamic intrinsic relations simultaneously varying from regions to regions~\cite{jacobs2016death,bettencourt2021introduction,rauws2020framework}.
Generally, urban factors form and can be categorized into three fundamental elements in urban science: \textit{Citizens},~\textit{Locations}, and~\textit{Mobility}.
First, the \textit{Mobility} refers to behaviors of citizens that would promote development in different urban locations.
Second, the \textit{Locations} would generate diverse routes among different places and thus to regulate movement patterns of~\textit{Mobility}.
Third,~\textit{Citizens}, as the most essential element involved with urban dynamics, would demand for urban infrastructures while traveling in the urban space to visit different locations, creating urban mobility, consumption, and social communication.
Nevertheless, there still lacks a fine-grained and fully penetrative comprehension on factor-level relations and interactions to obtain an in-depth understanding of urbanization from the perspectives of basic urban elements.
Given the wide availability of urban data from various domains and with the rising of artificial intelligence techniques,
we are inspired to uncover the highly intricate relationships among urban factors.
The discovered knowledge from both quantitative and qualitative perspectives would help counterparts to better understand urban development and promote insightful decisions for sustainable urban governance and resource management.

In the literature of urban science,
most existing works rely on correlation analysis to study the relationships among different urban factors~\cite{lu2018inferring,shi2021attentional,yue2017measurements,xu2020deconstructing}.
For instance, it has been revealed that the number of sports venues has a positive linear correlation with the number of local antidepressant prescriptions within the same neighborhood~\cite{hasthanasombat2019understanding}.
Nevertheless, as urban factors distinguish from each other in a region-by-region manner,
the relationships among them become vastly intricate in modern cities.
In particular, there exist ubiquitous confounding effects in urban space.
For instance, under the influence of an additional confounding factor,
two urban factors exhibit inverse correlations despite their actual relationship~\cite{pearl2000models}.
Recall the example, when the green space is introduced into the case as a confounding factor, it would not only constrain the construction of sports venues but also reduce local antidepressant prescriptions by creating a more psychologically-friendly environment.
As such, the relationship between the two factors in the previous example is indeed negative-correlated, which overturns the result obtained from correlation analysis.
In this study, we are motivated to precisely characterize the real relationships among urban factors by taming the universal confounding effect in urban space.
The key idea is to first discover causalities among urban factors and recover causal relations to determine how urban factors would directly or indirectly affect each other.
On the dimension level, we aim to achieve inference on more sophisticated causalities among citizens, locations, and urban mobility behaviors.
To this end, there are two primary challenges to our goals.
First, the solution space is enormous for the potential causal relations among urban factors. Second, it is quite daunting to eliminate the impact of confounding factors and discover authentic causal relations.


To comprehensively study the causal relations embedded among urban factors, we propose an urban causal analytic framework to tackle the potential deficiency in conventional correlation analysis.
By observing causal relations from an open-source urban dataset, we collect high-quality urban factors that depict citizens' demographic information, distribution of locations, and mobility behaviors across over 2,000 census tracts in the city of New York.
While most causal inference tasks require prior knowledge or assign typical causal relations based on common understandings, the actual relations embedded in numerous urban factors are intricate and thus hard to obtain. Therefore, inspired by the Reinforcement Learning (RL) method that can optimize decisions from an enormous searching space~\cite{zhu2019causal},
we first design a casual discovery algorithm to recover potential causal relations among urban factors.
The obtained causal graph further serves as the prior knowledge for both estimating causal effects and shaping comprehensive relations among categorized urban factors under elements of citizens, locations and mobility.
In addition, we leverage the propensity score matching method~\cite{caliendo2008some,10.1145/3365677,10.1145/3533725,10.1145/3444944} to alleviate confounding effects in the causal graph and characterize unbiased urban causal effects.
As a result, the sign and magnitude of estimated causal effects could precisely reveal the complicated interacting mechanism among urban factors.
Based on these observations, we further leverage the significance of estimated causal effects as a core criterion to select urban factors in the prediction task of mobility behaviors.


The causal analytic framework yields insightful results that shed light on discovering semantic relations among factors across basic urban elements from a causal perspective. First, we recover a three-tier hierarchical causal structure using the causal discovery approach.
In specific, citizens' demographics are at the top level, influencing both urban locations and mobility behaviors.
Meanwhile, location distributions lie at the middle level, casting causal effects on mobility behaviors. 
The causal graph provides structural and semantic urban knowledge while revealing confounding effects that negate simple correlation analysis.
Second, causal effects estimated from the causal graph also uncover the exact directions of causalities in urban space.
For example, the higher median age in a region would lead to less mobility flow, even if the observed correlation is positive.
Take another case, more elder residents would cause fewer local education venues regardless of their positive correlation.
In addition, education venues have a negative causal effect on the ratio of people taking public transportation.
The above deconfounded directions in causal inference can be applied to instruct interventions of urban governance.
Third, pruning factors that generate insignificant causal effects can remove more redundant information compared with selecting factors with significant correlations, thus enhancing the performance of urban mobility behavior prediction.
The improvements in urban mobility prediction tasks brought by causal analysis
validate the applicability of causal discovery and inference to the fields of social science
and urban computing.

The main contribution of this paper to can be summarized as follows.
\begin{itemize}
    \item We are the first to comprehensively analyze causal relations among citizens, locations, and mobility behaviors from observational urban factor data. The recovered causal graph reveals a three-tier hierarchical causal structure in urban space, where citizens are the primary cause of location distributions and mobility behaviors.
    \item We provide quantitative computations and qualitative analysis on causal effects among urban factors by applying causal inference methods to reduce confounding effects. The estimated causal effects reflect complex relationships among numerous urban factors under dimensions of citizens, locations, and mobility. The directions of estimated causal effects also provide comprehensive insights into the relations among urban factors. 
    \item We further apply the estimated causal effects into urban mobility prediction tasks. Extensive experimental results demonstrate that after pruning factors with insignificant causal effects, urban mobility prediction models can achieve better and more robust performance. This sheds light on leveraging causal effects as a metric for selecting significant factors for urban computing tasks.
\end{itemize}

The rest of this paper is organized as follows. We first introduce the urban dataset and formulate the research problem in Section~\ref{section:data}. Section~\ref{section:method} presents the causal analytic framework, which consists of a causal discovery method and the causal effect estimation approach. Results and analyses are presented in Section~\ref{section:results}. Related works and implications of our work are discussed in Section~\ref{section:discussion}.

\section{Motivation and Problem Overview} \label{section:data}
\begin{figure*}[t]
  \centering
      \begin{subfigure}[Correlation table.]{
      \label{gra:corr}
        \includegraphics[height=0.38\textwidth]{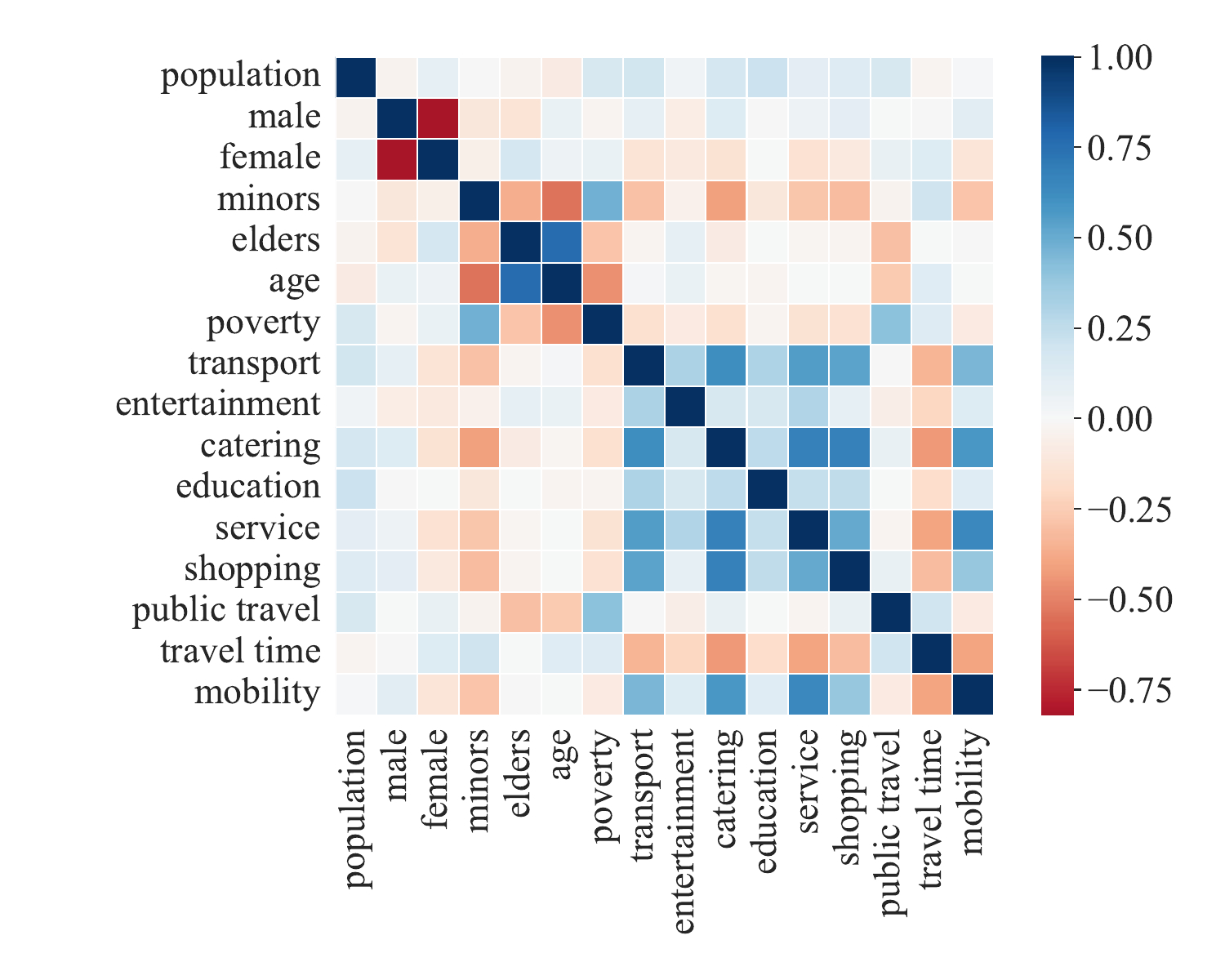}}
      \end{subfigure}
      \begin{subfigure}[Factor distributions.]{
      \label{gra:dis}
        \includegraphics[height=0.38\textwidth]{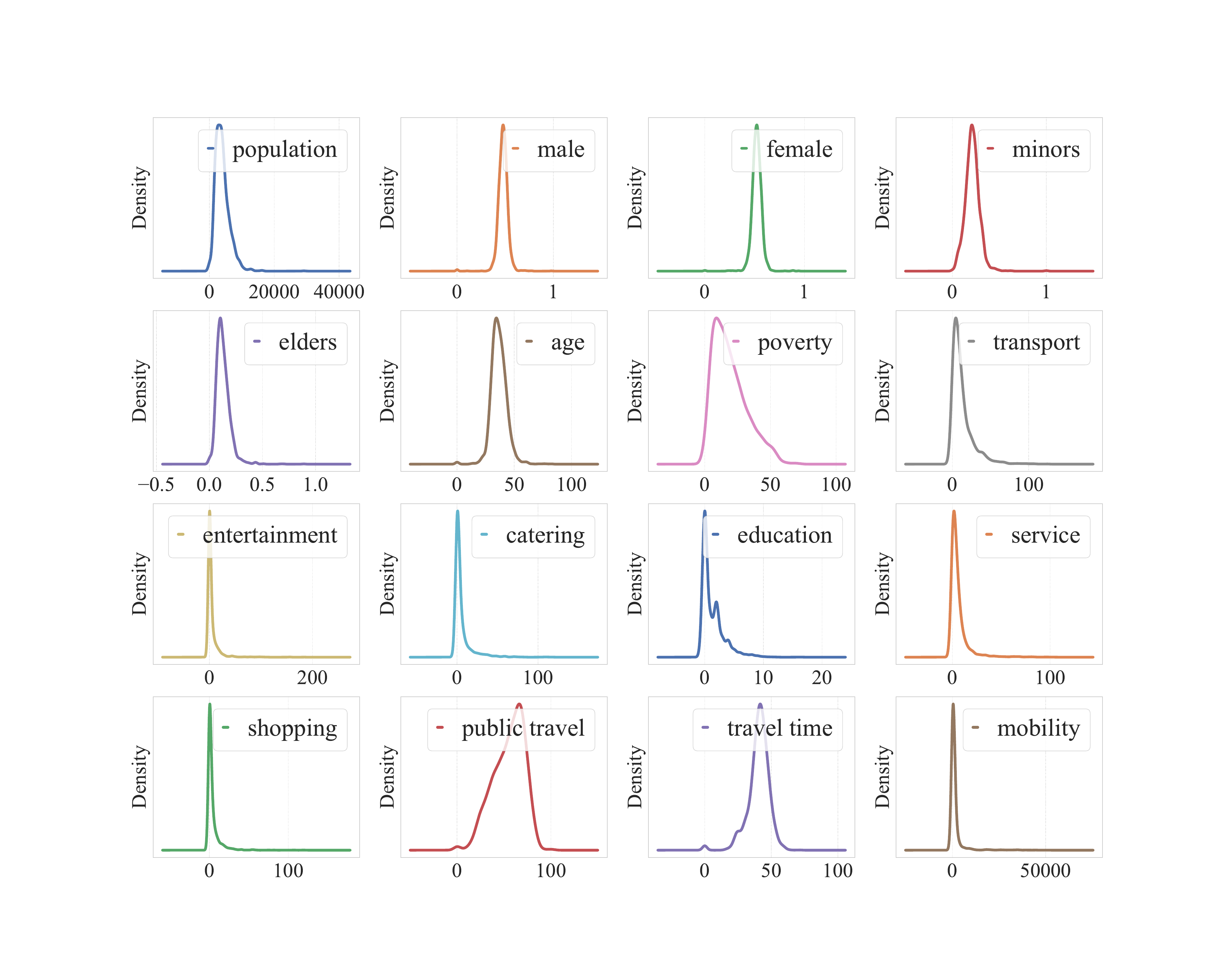}}
      \end{subfigure}
      \caption{Statistics of the urban dataset.}
\end{figure*}

We formulate the research problem as discovering and inferring the causal relationships and effects among citizens, locations and mobility, which is motivated by the deficiency of correlation analysis in complex urban interactions. 
Studying the relationship between each pair of factors from citizens, locations, and mobility can be beneficial to various practical applications,
such as urban demand forecasting and facility planning~\cite{liu2017point,kadar2018mining,liu2017point,kadar2018mining,10.1145/3356584,10.1145/3340847}.
However, correlation-based analysis~\cite{lu2018inferring,shi2021attentional,yue2017measurements,10.1145/3502738,10.1145/1497577.1497578} may easily lead to contrary conclusions to the common sense, due to the widely existed confounding effects~\cite{vanderweele2013definition,mcnamee2003confounding}.
For example, Figure~\ref{gra:corr} shows a correlation diagram between each pair of urban factors.
As it reveals, urban factors of median age and population mobility are positively correlated, \ie, people living in areas with a high degree of aging would have more active mobility. This is a conclusion that contradicts our common sense~\cite{metz2000mobility,macpherson2007health}, which is caused by the confounding effect. 

Causal inference~\cite{heckerman1999bayesian,spirtes2016causal} has been widely used in many fields to tame biases in relationship estimation caused by the confounding factors. Conventional causal inference methods~\cite{zheleva2021causal,imbens2015causal} tend to rely on assumptions and make estimations based on prior knowledge acquired from causal graphs between pair-wise factors.
Nevertheless, it is quite difficult to directly obtain the causal graph among various urban factors through prior knowledge.
The key reason is that searching such a graph is an NP-hard problem~\cite{spirtes2016causal,malinsky2018causal},
and the prior knowledge cannot fully include relationships between all pairs of factors.
Therefore, we first design a reinforcement learning-based causal discovery method to obtain the causal graph between
numerous urban factors.
The RL-based causal discovery method models the causal graph searching process as a sequential decision problem,
so as to reduce both the searching space and the difficulty of NP-hardness.
The obtained causal graph is further applied to discover the causal relationship and estimate the causal effect among urban factors in citizens, locations, and mobility.
The discovered causal relationships can facilitate downstream tasks related to urban computing (\ie, mobility prediction and urban planning).

\section{Method} \label{section:method}
\subsection{Overview}
\begin{figure*}[t]
\centering
\begin{subfigure}{
\includegraphics[width=0.8\textwidth]{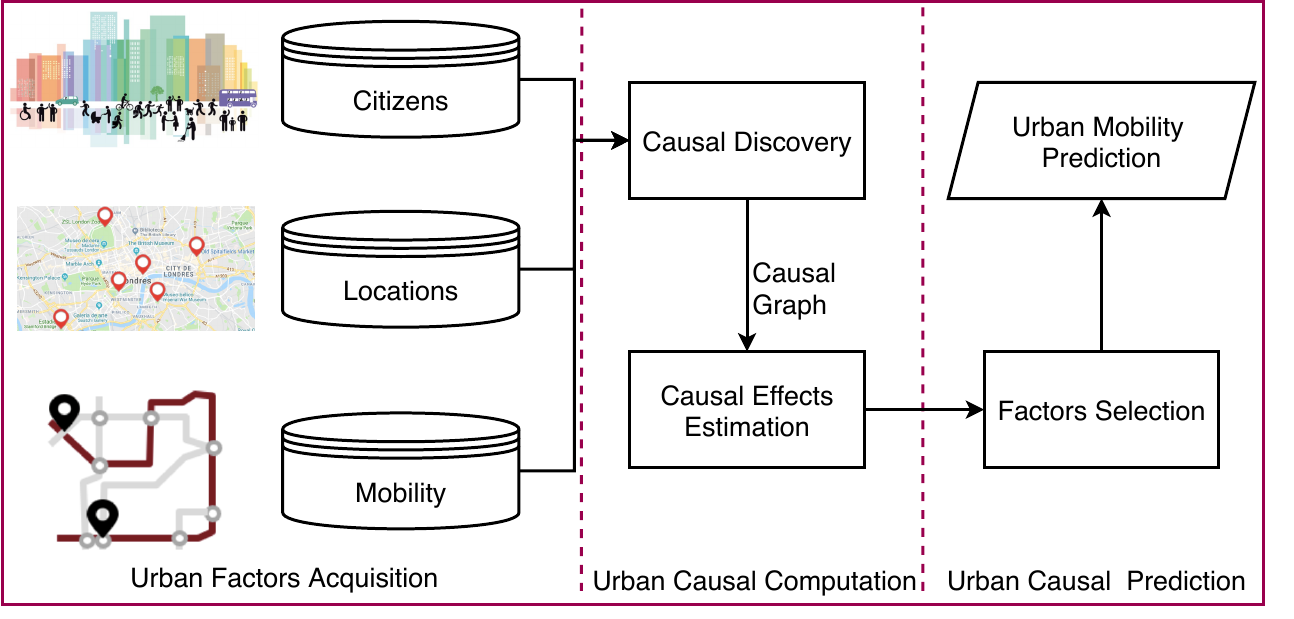}}
\end{subfigure}
\caption{The framework of our method.}
\label{fig:frame}
\end{figure*}

Fig.~\ref{fig:frame} provides an overview of the proposed framework for discovering causal relations among urban factors across fundamental urban elements.
Overall, it consists of the following three basic components.
1) \textbf{Urban Factors Acquisition}. We first introduce the dataset used in this work and pre-process data to obtain a variety of factors from citizens, locations and mobility.
2) \textbf{Urban Causal Computation}. Second, we propose a reinforcement learning-based causal discovery algorithm to obtain the causal graph and depict the causal relations among urban factors. To model the intensity of acquired causal relations, we estimate the causal effects between causality-related factors with a matching method to alleviate confounding effects in the causal graph.
3) \textbf{Urban Causal Prediction.} To evaluate the effectiveness of acquired causal relations and effects, we predict urban mobility factors with significant cause factors derived from the causal graph. 

\begin{table*}[t]
\centering
\caption{The basic information of the dataset.}
\begin{tabular}{c|c|c|c|c}
\hline
\textbf{City} & \textbf{Period}     & \textbf{Data Type} & \textbf{Region Number} & \textbf{Region Level} \\ \hline
 New York       & 2019 $\sim$ 2020 & Citizens, Locations, Mobility       & 2,167   & Census Tract                    \\ \hline
\end{tabular}
\label{tab:info}
\end{table*}

\begin{table*}[t]
\centering
    \caption{The composition of citizens, locations and mobility.}\label{table:data2}
    \resizebox{4.in}{!}{
    \begin{tabular}{cl}
        \hline
         \textbf{Data Type} & \multicolumn{1}{c}{\textbf{Factors}}\\\hline
         \textbf{\emph{Citizens}}     & \begin{tabular}[c]{@{}l@{}}
        Total population, Male rate, Female rate, \\Minors rate,
Elders rate\\ Median age,  Poverty level
\end{tabular}\\\hline
        \textbf{\emph{Locations}}   & \begin{tabular}[c]{@{}l@{}}
         Transport, Entertainment,\\
        Catering, Education \\
        Service,
        Shopping\end{tabular}\\\hline
        \textbf{\emph{Mobility}}   & \begin{tabular}[c]{@{}l@{}}
       Proportion of people traveling by public transport \\
        Mean travel time to work\\
        Population mobility\end{tabular}\\\hline
    \end{tabular}}
\label{tab:con}
\end{table*} 

\subsection{Urban Factors Acquisition}
The utilized dataset is collected from New York city, the U.S.
As shown in Table~\ref{tab:info}, the dataset depicts urban factors categorized by Citizens, Locations and Mobility across 2,167 census tract.
Specifically, there exist 16 types of urban factors, whose distribution is shown in Figure~\ref{gra:dis}.
To study the causal relationships among these factors, we refer to the literature on the urban relations and categorize them into three dimensions in Table~\ref{tab:con}.
In addition, we acquire source data for different urban factors as follows.
First, source data for factors of \textit{Citizens} are collected from the United States Census Bureau~\cite{USCB}.
Second, source data for factors of \textit{Locations} are crawled from the OpenStreetMap (OSM)~\cite{OpenStreetMap}, which is a crowd-sourced map data collection service.
Third, source data for factors of \textit{Mobility} are collected from the Origin-Destination Employment Statistics (ODES) organized by the Longitudinal Employer-Household Dynamics (LEHD)~\cite{USCB} program of the United States Census Bureau.
The above data sources together make the dataset sufficient for discovering and inferring the causal relationships among basic urban elements. Next, we introduce in detail how we design factors to depict citizens, locations, and mobility from the collected dataset.


\textbf{Citizens} depicts the basic factors of the population demography, such as gender, age, number, \etc.
Existing works~\cite{liu2017point,shimosaka2015forecasting,kadar2018mining} have studied the relationship between urban factors in \textit{Citizens} and \textit{Locations} distribution.
It suggests that  the demand for Locations in a region will be affected by the citizens living in the same region, and the Locations distribution in a region is closely related to the gender ratio and poverty rate of the population through modeling.
Some other studies\cite{wang2017region,xu2020sume} exploited factors of citizens to predict the mobility of the region, finding that the population and age distribution of each region are important factors to influence urban mobility. Based on the above knowledge and the collected dataset, we select the total population, male rate, and female rate to depict urban population and gender factors.
Moreover, minors rate, elders rate and median age are selected to depict age factors.
At last, the poverty factor is aligned with the poverty level indicator from the collected dataset.

\textbf{Locations} mainly depicts the number of different types of locations in the region. Previous studies~\cite{liu2017point,shimosaka2015forecasting,kadar2018mining,zeng2017visualizing,chen2020learning,yao2018exploiting} have explored the relationship among locations, citizens and mobility.
The key finding is that the distribution of locations can influence regional indicators, such as poverty level~\cite{zeng2017visualizing,chen2020learning,yao2018exploiting}.
Therefore, we select 6 types of locations as the urban factors, including transport,
entertainment, catering, education, service, and shopping. The total numbers of each location factors are used as indicators to study the influence of urban locations.

\textbf{Mobility} includes the mobility of the population and factors related to mobility.
Previous studies~\cite{zhang2020real,fang2019mac,ruan2020dynamic,zhao2019celltrans} have shown people's travel choices and travel time are important factors of mobility.
Therefore, we take the proportion of people traveling by public transport, mean travel time to work, and population mobility as the key factors in our research.

\subsection{Urban Causal Computation}

\textbf{Causal Discovery via Deep Reinforcement Learning.}
\label{section:discovery}
Searching for the causal graph among multiple urban factors is an NP-hard combinatorial optimization problem~\cite{spirtes2016causal,malinsky2018causal},
which is difficult to solve by conventional causal discovery methods.
Reinforcement learning is powerful for solving such large-scale combinatorial optimization problems~\cite{mazyavkina2021reinforcement,barrett2020exploratory,bello2016neural}.
Therefore, we design a RL-based causal discovery method to search the causal graph among urban factors. We first formulate the problem as a one-step Markovian Decision-making Process (MDP). Formally, each MDP can be described as a 4-tuple $\left (\mathit{S},\mathit{A},\mathit{P},\mathit{R} \right )$. Specially, $\mathit{S}$ and $\mathit{A}$ represent the state space and action space, respectively. Moreover, $\mathit{P}:\mathit{S}\times \mathit{A}\rightarrow \mathbb{R}$ represents the probability of state transition. That is, $\mathit{P(s_{t+1}|s_{t},a_{t})}$ is the probability distribution of the next state $s_{t+1}$ conditioned on the current state $s_{t}$ and action $a_{t}$. Finally, $\mathit{R}:\mathit{S}\times \mathit{A}\rightarrow \mathbb{R}$ is the reward function with $\mathit{R(s,a)}$ representing the reward received by executing action $a \in \mathit{A}$ under the state $s \in \mathit{S}$. In addition, for the convenience of modeling, we indicate the full set of factors using $\mathcal{F} = \{ F_i | i = 1,2,...,|\mathcal{F}| \}$, where  $|\mathcal{F}|=16$ in our dataset. In the following, we detail how to model the above components of the MDP.

\begin{itemize}
    \item \textbf{State:} 
    As mentioned in previous studies~\cite{wang2021ordering,zhu2019causal}, it is difficult to capture the underlying causal relationships by directly using the urban factors $F$ as the state. These studies also inspire us to use an encoder to embed each factor $F_{i}$ to state $s_{i}$, which is beneficial for the causal discovery process.
    Therefore, the state space can be obtained as $\mathit{S}=(s_{1},s_{2}...,s_{|\mathcal{F}|})=encoder(F_{1},F_{2}...,F_{|\mathcal{F}|})$.
    Motivated by~\cite{wang2021ordering}, we exploit a self-attention based encoder in the Transformer structure in the proposed reinforcement learning model.

    \item \textbf{Action:}  
     The action of our RL framework is to generate a binary adjacency matrix $U$, which corresponds to the causal graph $\mathcal{G}$. For example, if the $i$-th row and $j$-th column of $U$ is 1, it means that the $i$-th factor points to the $j$-th factor in the causal graph $\mathcal{G}$.
     
     \item \textbf{State transition:} 
    In the one-step MDP, the state $s$ will be directly transferred to the end state of the episode after the first action is executed.

    \item \textbf{Reward:} 
     Our optimization goal is to search over the space of all Directed Acyclic Graph (DAG) to find  a $\mathcal{G}$ with the minimum Bayesian Information Criterion score $S_{BIC}(\mathcal{G})$ \cite{burnham2004multimodel,kuha2004aic}, which depicts how well the obtained DAG causally matches the observed data. Here, we adopt the DAG constraint $\rho(\mathcal{G})$ proposed by~\cite{zheng2018dags} as a penalty term to add to our optimization goal, thus to make sure that the $\mathcal{G}$ is a DAG. Therefore, we set the episode reward as $R=-S_{BIC}(\mathcal{G})-\rho(\mathcal{G})$, which will be obtained when executing the binary adjacency matrix $U$.
     The process of RL learning will maximize reward $R$, thereby minimizing $S_{BIC}(\mathcal{G})$ and $\rho(\mathcal{G})$ simultaneously.
\end{itemize}

\begin{figure*}[h]
  \centering
      \begin{subfigure}{
        \includegraphics[width=0.97\textwidth]{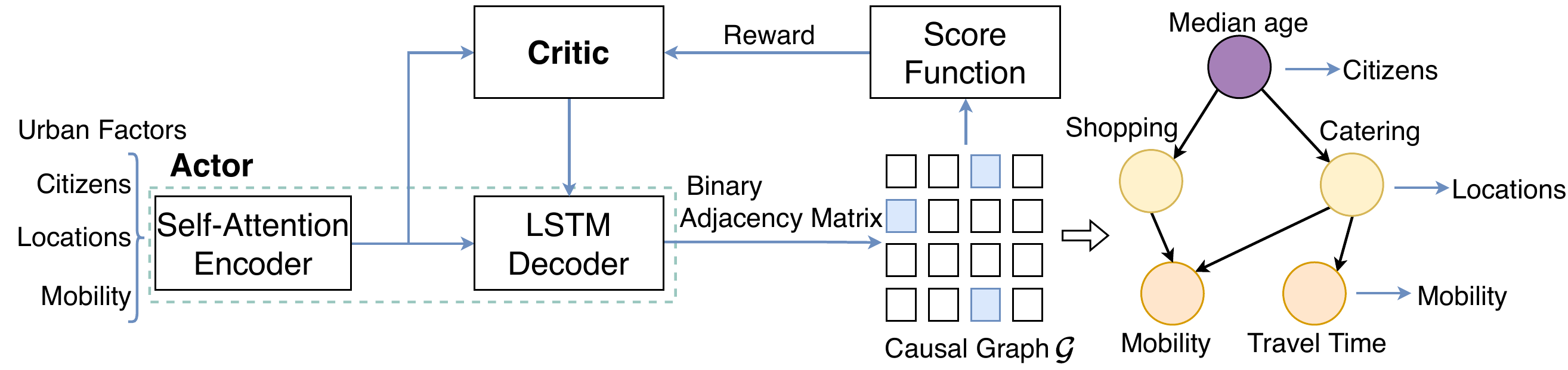}}
      \end{subfigure}
      \caption{The framework of  causal discovery method via deep reinforcement learning.}
  \label{fig:RL}
\end{figure*}

Based on the above MDP models, the  causal discovery is described by a policy function $\pi :\mathit{S} \rightarrow \mathit{A}$. Specially, $\pi(a|s)$ represents the probability of choosing action $a$ under current state $s$. The framework of causal discovery method is shown in Figure \ref{fig:RL}. We adopt a self-attention encoder and an LSTM based decoder~\cite{zhu2019causal} to map the state to action. In addition, we introduce the actor-critic algorithm~\cite{konda2000actor} to train the proposed RL framework to obtain the best causal graph $\mathcal{G}$.
Then, the nodes and edges of graph $\mathcal{G}$ depict the urban factors and their causal relationships, respectively, as shown in Figure~\ref{fig:RL}.

\textbf{Estimating Causal Effects from Causal Graph}.~\label{section:inference}
Edges in the causal graph mark confident causal relations among regional factors in the urban space. An unresolved issue is to measure the extent and direction of causalities from source nodes to target nodes, which ultimately shapes the understanding of fundamental functioning mechanisms in urban space. Meanwhile, a crucial challenge for obtaining accurate causal effects is that observed correlation between causal and effected factors are potentially confounded by a confounding factor.
The confounding factor simultaneously exerts causal effects on both the causal and the effected nodes, resulting in the well-known confounding effect~\cite{vanderweele2013definition}.
To solve this challenge, we leverage a propensity score-based method that controls the distribution of confounding variables over different causal factors to alleviate the confounding effect.
Under balanced distributions of confounding variables, we are able to identify debiased causal effects of each edge in the estimated causal graph.

In the learned causal graph, source nodes and target nodes are called \emph{treatments} and \emph{outcomes}, respectively. Here, we adopt the \emph{potential outcome} framework by Rubin~\cite{rubin1974estimating}. The \emph{potential outcome} is the expected value of the \emph{outcome} factor when the \emph{treatment} factor is set to a certain level. The causal effect, also known as the \emph{treatment effect}, is the difference in \emph{potential outcomes} under different treatment levels. This measurement requires \emph{counterfactual} outcomes as we can only observe one outcome. A traditional approach to obtain the \emph{counterfactual} outcomes and estimate treatment effects is the Randomized Controlled Trial (RCT). In an RCT, units are randomly assigned with a level of \emph{treatment} regardless of their other attributes. Therefore, these attributes are distributed identically in each treatment group, which ensures that groups are comparable to each other's counterfactual outcomes. 

\begin{figure*}[]
  \centering
      \begin{subfigure}[Confounding effect.]{
      \label{fig:confounding}
        \includegraphics[height=0.25\textwidth]{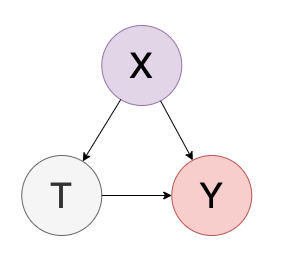}}
      \end{subfigure}
      \begin{subfigure}[Calculating propensity score.]{
      \label{fig:prop}
        \includegraphics[height=0.25\textwidth]{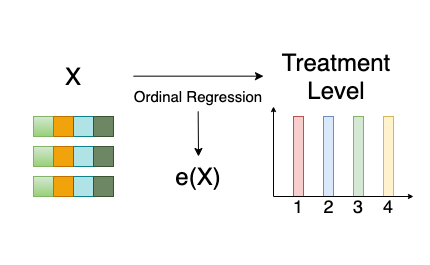}}
      \end{subfigure}
      \begin{subfigure}[Matching process.]{
      \label{fig:match}
        \includegraphics[height=0.25\textwidth]{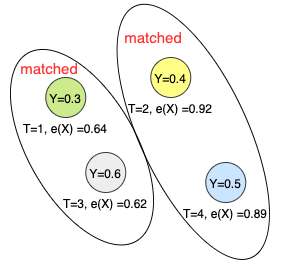}}
      \end{subfigure}
      \caption{The framework of estimating causal effects.}
      \label{fig:psm}
\end{figure*}

However, randomized controlled trials are impractical when the observed data is confounded. Confounding factors are factors that simultaneously affect both nodes on a causal edge.
For instance, the node $X$ in Figure~\ref{fig:confounding} affects both nodes $T$ and $Y$. Therefore, we adopt the propensity score matching strategy~\cite{caliendo2008some} to alleviate the confounding effects during estimating causal effects in the proposed causal graph.
\emph{Matching} is a widely used method to reproduce a randomized trial, where units with different treatment levels but exact same confounding factors are matched to form a counterfactual pair. When there are multiple confounding factors, matching units with identical confounding factors might be impossible. The propensity score, as a numeric value, is introduced to compensate for this deficiency. When matching units with similar propensity scores, we can obtain balanced confounding factors.
The propensity score $e(X)$ is a scalar determined by confounding factors $X$.
It is expected to be a balancing score that the conditional distribution of $X$ given $e(X)$ is the same for all treatment level $T$~\cite{rosenbaum1983central},
\begin{center}
\begin{equation}
    X \perp T | e(X).
\label{equ:balancingscore}
\end{equation}
\end{center}
Note that matching units with similar $e(X)$ instead of high-dimensional confounding factors should still balance the confounding factors. 

Urban region factors are continuous and multilevel instead of binary.
In this work, we follow the design of propensity score for multiple treatment levels introduced in~\cite{hasthanasombat2019understanding}. With each urban region factor as the treatment factor, we first group all census tracts into four equally sized quantile groups based on their values on that factor. Then, we use an ordinal regression model to fit the ordinal treatment level by confounding factors and derive propensity scores.
The ordinal regression model has the form of
\begin{equation}
\centering
\label{equ:ord}
    P(T\le d | X) = \sigma(\theta_d - \textbf{w}^TX),
\end{equation}
where $T$ is the treatment level, $d$ is a level ranging from the lowest to the highest quantile, $\textbf{w}$ and $\theta_d$ are parameters to learn, and $\sigma(\cdot)$ is the logistic function $\sigma(x)=\frac{x}{1+e^{-x}}$. Let $e(X) = \textbf{w}^TX$, then $P(T|X)$ only depends on $e(X)$, which implies that the treatment level and confounding factors are independent conditioned on $e(X)$. Note that $e(X)$ satisfies the characteristic of balancing score in Eq.~\ref{equ:balancingscore}. Therefore, for each treatment and its confounding factors, we fit the ordinal regression model and use learned model parameters to estimate the propensity score $\hat{e}(X)=\hat{\textbf{w}}^TX$. The procedure of fitting an ordinal regression model to calculate propensity score is shown in Figure~\ref{fig:prop}.

Then we match census tracts with similar propensity scores to alleviate the confounding effects. We first define the distance between two census tracts based on their treatment level and estimated propensity scores.
We adopt the distance defined in~\cite{lu2001matching}, which indicates the difference in estimated propensity score divided by the difference in treatment level as follows.
\begin{equation}
\centering
    d_{i,j} = \frac{|\hat{e}(X_i) - \hat{e}(X_j)|}{|T_i - T_j|},
\end{equation}
where $d_{i,j}$ denotes the distance between the $i$-th and $j$-th census tract.
Under this metric, census tracts with similar propensity scores and dissimilar treatment levels will become closer.

In the matching procedure, each census tract is matched with its closest census tract, which will serve as the counterfactual outcome. For instance, in Figure~\ref{fig:match}, the green region is matched with the grey region, while the yellow region is matched with the blue one. Then, for each pair of census tracts, we calculate the estimated individual treatment effect, by using the difference in outcome divided by the difference in treatment level. The Average Treatment Effect (ATE) from the treatment factor to the outcome factor is estimated as the average value of all individual treatment effects. As a result, the value of ATE represents the change in outcome factor when the treatment is raised by one level. In addition, we can also analyze the significance of an ATE by testing if the individual treatment effects have a zero average by statistical tests. If the ATE deviates from 0 with a small p-value, the corresponding causal relation is then regarded as significant.

\subsection{Urban Causal Prediction}
Based on the significance tests on estimated ATEs, we propose a causality-based method for the input selection stage of urban supervised learning tasks. A common issue in supervised learning is that not all factors input to the prediction model contain sufficient information of the outcome. For example, model inputs with a large variance but insignificant impact on the outcome might be falsely perceived by models in the training stage, deteriorating their performances on test sets. Moreover, most of the prediction models assume that training samples and test samples are identically distributed. However, if this assumption fails, correlations in training samples might not hold in the test set. Models that mistakenly learn these correlations might have worse performance on the test set. To solve the above challenges, we suggest using the learned causal effects in selecting model input for the prediction of urban mobility behavior factors~\cite{zhang2021quantifying}.

The causal-based input selection procedure first follows the proposed urban causal computation process to estimate the causal effects between urban factors on the training set. Causal edges with significance level \emph{p > 0.05} are then pruned from the potential graph. For the prediction of a specific outcome factor, we only keep the factors located on any path in the pruned graph, which targets the outcome factor as the prediction model's inputs. Therefore, there are significant causal effects that flow from the model inputs to the outcome factor. 

\section{Results and Analysis} \label{section:results}

\begin{table*}[t]
	\caption{The causal order of urban factors in NYC.}
	\label{tab::performance}
	\centering
	\scalebox{1}{
		\begin{tabular}{ccc}
			\toprule[1pt]
			\bf Level &\bf Factors
			\\ \hline \hline
			\multirow{3}{*}{Top}    
			  &  Poverty level \\  
			     &   Minors rate
     \\  
			     &  Total population 
      \\
			\hline
			\multirow{3}{*}{Middle}    
 			      &  Catering \\  
			     &  Shopping   \\   
			    &   Service   \\  
			\hline
			\multirow{3}{*}{Bottom} 
			    & Proportion of people traveling by public transport \\
			   & Mean travel time to work \\   
			      &  Population mobility\\  
			\hline
			\bottomrule[1pt]
	\end{tabular}}
\label{table:order}
\end{table*}

\begin{figure}[t]
\centering
\subfigure[]{\label{fig:sub_g_a} \includegraphics[height=0.25\textwidth]{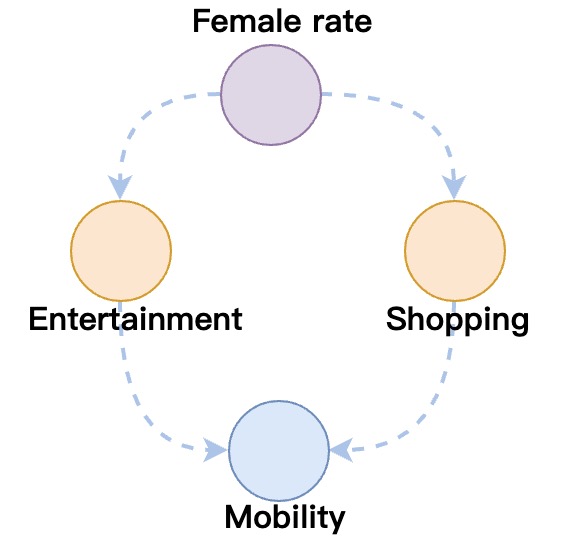}}
\subfigure[]{\label{fig:sub_g_b}\includegraphics[height=0.25\textwidth]{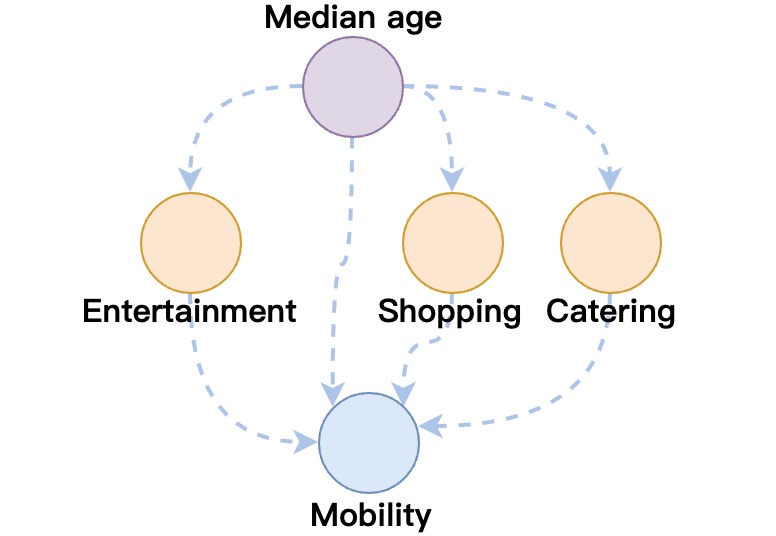}}
\subfigure[]{\label{fig:sub_g_c} \includegraphics[height=0.25\textwidth]{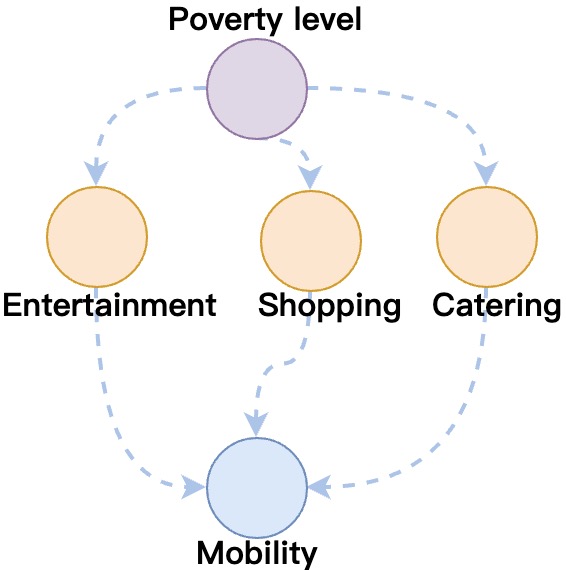}}
\subfigure[]{\label{fig:sub_g_d} \includegraphics[height=0.25\textwidth]{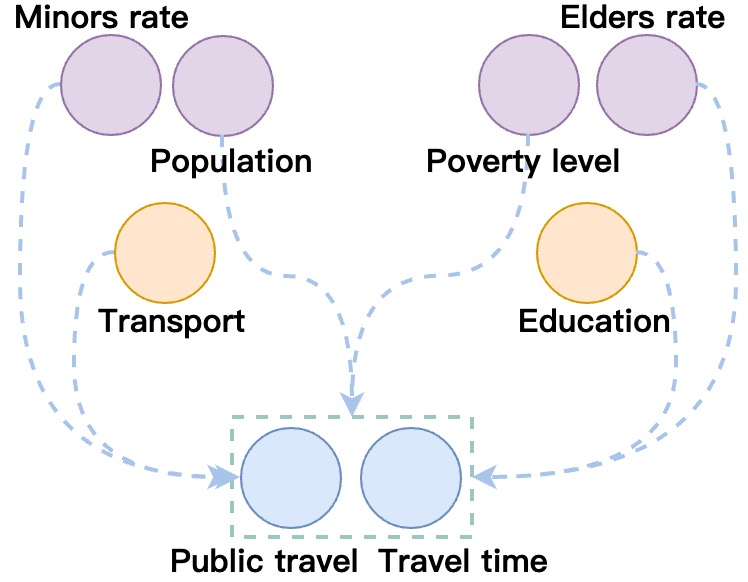}}
\caption{Four sub-graphs of our discovered causal graphs.}
\label{fig:sub_g}
\end{figure}

\subsection{Urban Causal Graph}
Based on the DRL-based causal discovery method, we divide urban factors into three levels according to the causal order~\cite{wang2021ordering} of the causal graph, as shown in Table \ref{table:order}. From this table, we can find that {\em factors of citizens always appear as top causes, while location factors appear as the middle result of citizens as well as the cause for mobility}. In addition, factors in citizens can influence mobility in two ways, which profoundly reveals two major causes for the formation of mobility in the urban space.
On the one hand, as basic urban attributes, citizens' factors will directly affect the formation and development of urban mobility.
On the other hand, the citizens' factors can also affect the distribution of locations and indirectly change the attraction to urban mobility, thus influencing urban mobility patterns.
Besides, our discovery that factors in citizens are the cause of the locations distribution supports conclusions in the existing literature~\cite{weidlich1999fast,portugali2000self}.
In urban space, the distribution of locations will change rapidly as the citizens changes, but it is a long process for citizens to gradually influence factors in location dimension.
This is because factors in citizens are the cause of the locations distribution, which coincides with our findings.

To further analyze the causal relationship among urban factors in details,
we provide sub-graphs of discovered causal graph, as shown in Figure~\ref{fig:sub_g}.
From Figure~\ref{fig:sub_g_a}, we can find that {\em the female rate in a region is the cause for distribution of the shopping and entertainment locations, and it indirectly affects mobility through these locations}. This is because women tend to have stronger consumer  behaviors~\cite{herter2014man,philips2000shopping}, thus female rate will become an important cause for the location of shopping and entertainment locations.
With both the above locations attracting citizens,
the female rate will indirectly affect mobility factors in surrounding areas.
From Figure~\ref{fig:sub_g_b}, we can find that {\em the average age of the regional population directly and indirectly affects mobility through locations}. This is because people in different ages have different traveling abilities, which will directly affect the mobility of the area. In addition, people of different age groups will be attracted by different types of locations, therefore, the average age indirectly affects mobility based on the attraction of locations to mobility. 
From Figure~\ref{fig:sub_g_c}, we can find that {\em the poverty rate will indirectly affect mobility through locations}. This shows that the locations site selection process will fully consider the degree of wealth of the region. 
From Figure~\ref{fig:sub_g_d}, we can find that {\em multiple causes are affecting the proportion of people traveling by public transport and mean travel time to work}.
The proportion of people in different age groups will have an impact on the two results, which is caused by the {\em different travel preferences of people in different age groups}. For example, middle-aged people often prefer to take private cars while minors often prefer to take public transportation~\cite{nevelsteen2012controlling,cheng2019applying}.
In addition, the varying travel duration by different vehicles varies will also affect the above two factors.
We can also find that the total population and poverty level of a region can have an affect the above two factors, which shows that {\em people in areas with different incomes and different populations often have different travel preferences}.

\subsection{Estimated Causal Effects}
The three-tier hierarchical causal graph depicts the flow of causal relations in the urban space.
For instance, citizens affect the distribution of local locations,
and factors in both citizens and locations lead to changes in urban mobility factors (\ie, the mean travel time of residents, the proportion of people taking public transportation, and local mobility flow).
However, it only demonstrates the source and the target in causal relations, lacking the direction of causal effects.
In other words, the causal graph learned via the reinforcement learning method does not intuitively answer the counterfactual question of `if the region has more transportation locations, whether the mobility flow increases or decreases'. The direction of causal effect cannot be straightforwardly represented by the correlation coefficient between the `causal' factor and the `effect' factor, due to the confounding effect depicted in Figure~\ref{fig:confounding}. Therefore, we adopt the propensity score matching method to alleviate confounding effects for all confounded causal edges in the causal graph. 

\begin{figure*}[]
  \centering
      \begin{subfigure}[Causal graph learned from causal discovery algorithm. Blue squares denote potential causal relations.]{
      \label{fig:causal_graph_matrix}
        \includegraphics[width=0.45\textwidth]{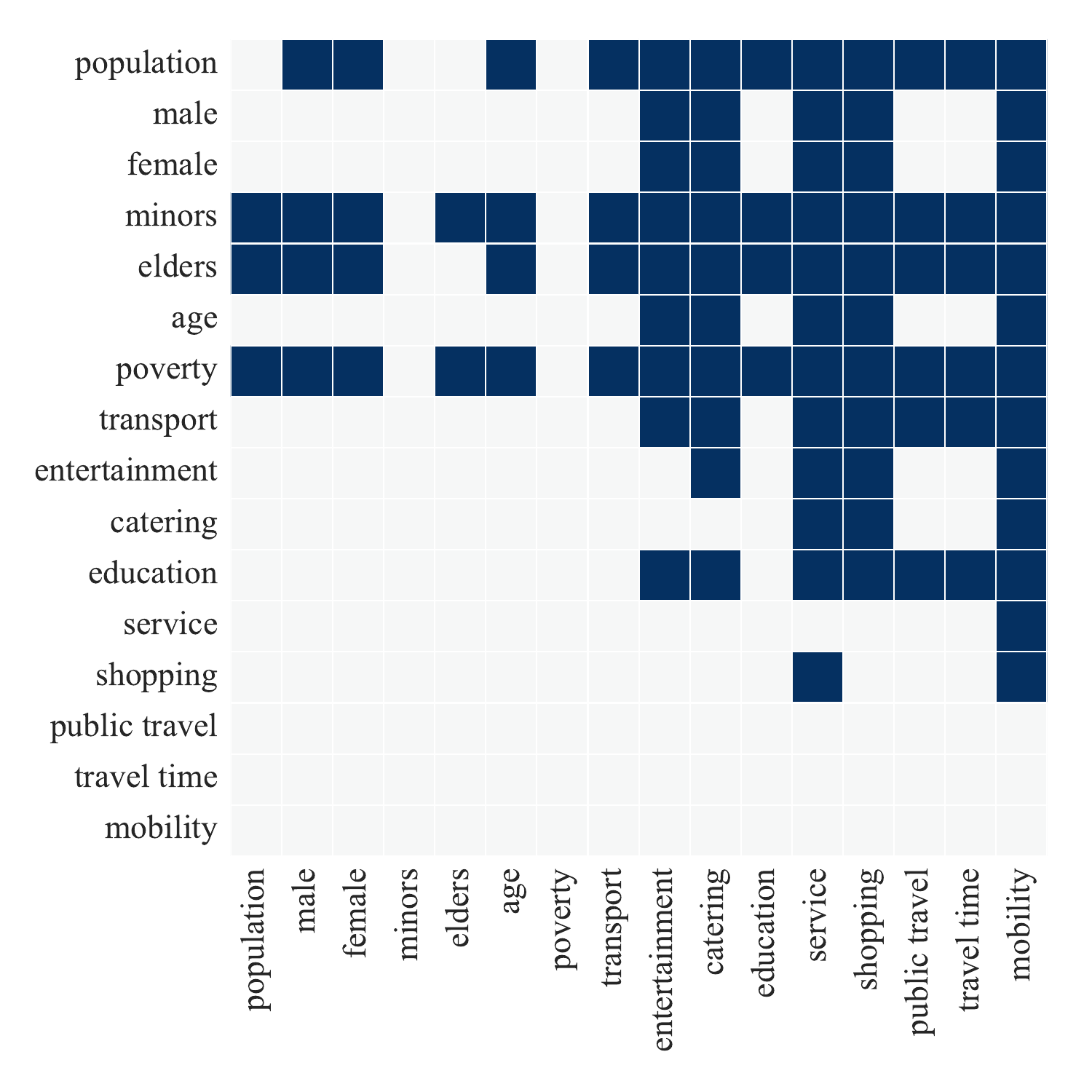}}
      \end{subfigure}
      \begin{subfigure}[Significant causal effects. Blue squares: positive effects. Red squares: negative effects.]{
      \label{fig:significant_causal}
        \includegraphics[width=0.45\textwidth]{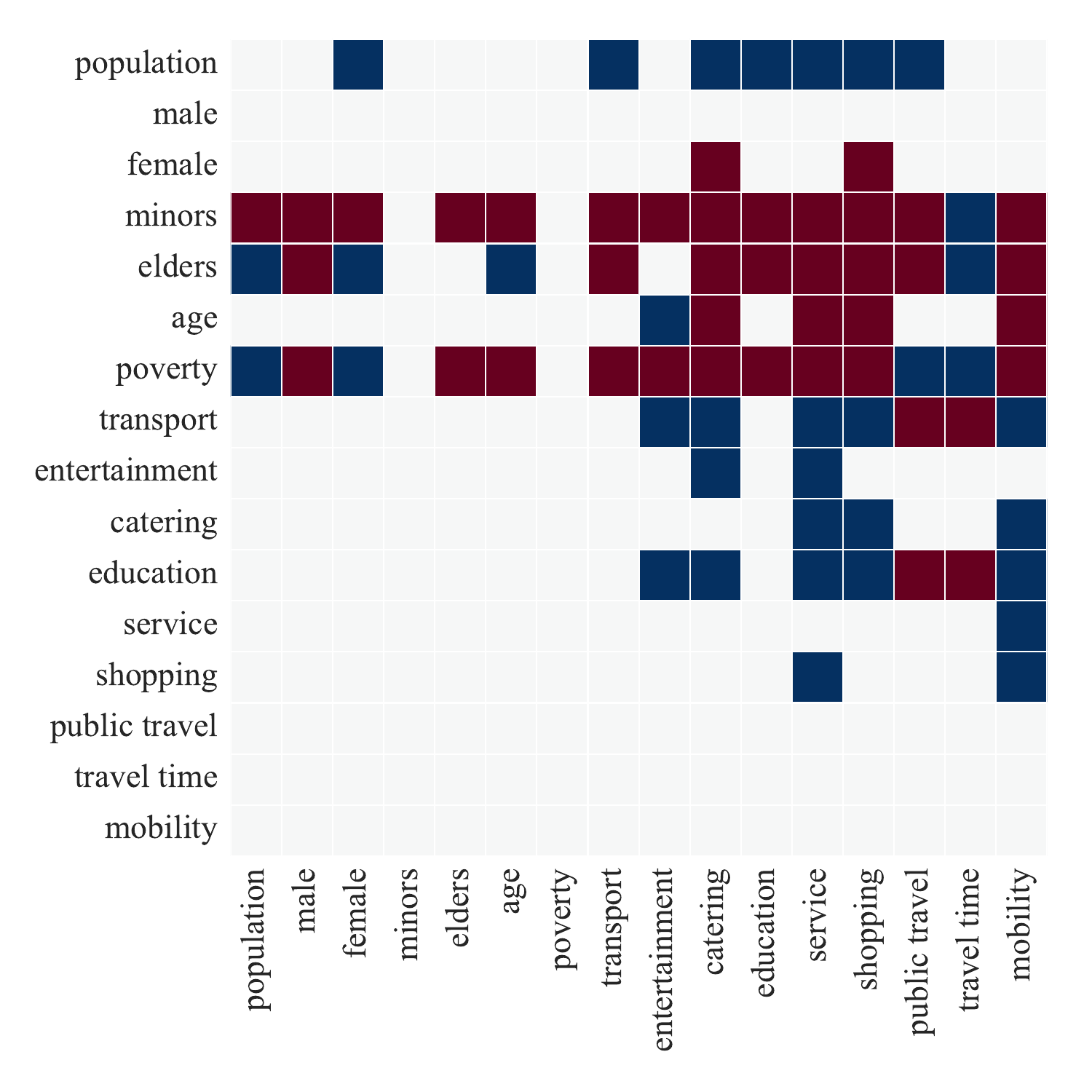}}
      \end{subfigure}
      \caption{Learned causal graph and significant causal effects after balancing confounding factors.}
\end{figure*}

\begin{figure*}[]
  \centering
      \begin{subfigure}[Confounding factors of population.]{
        \includegraphics[height=0.18\textwidth]{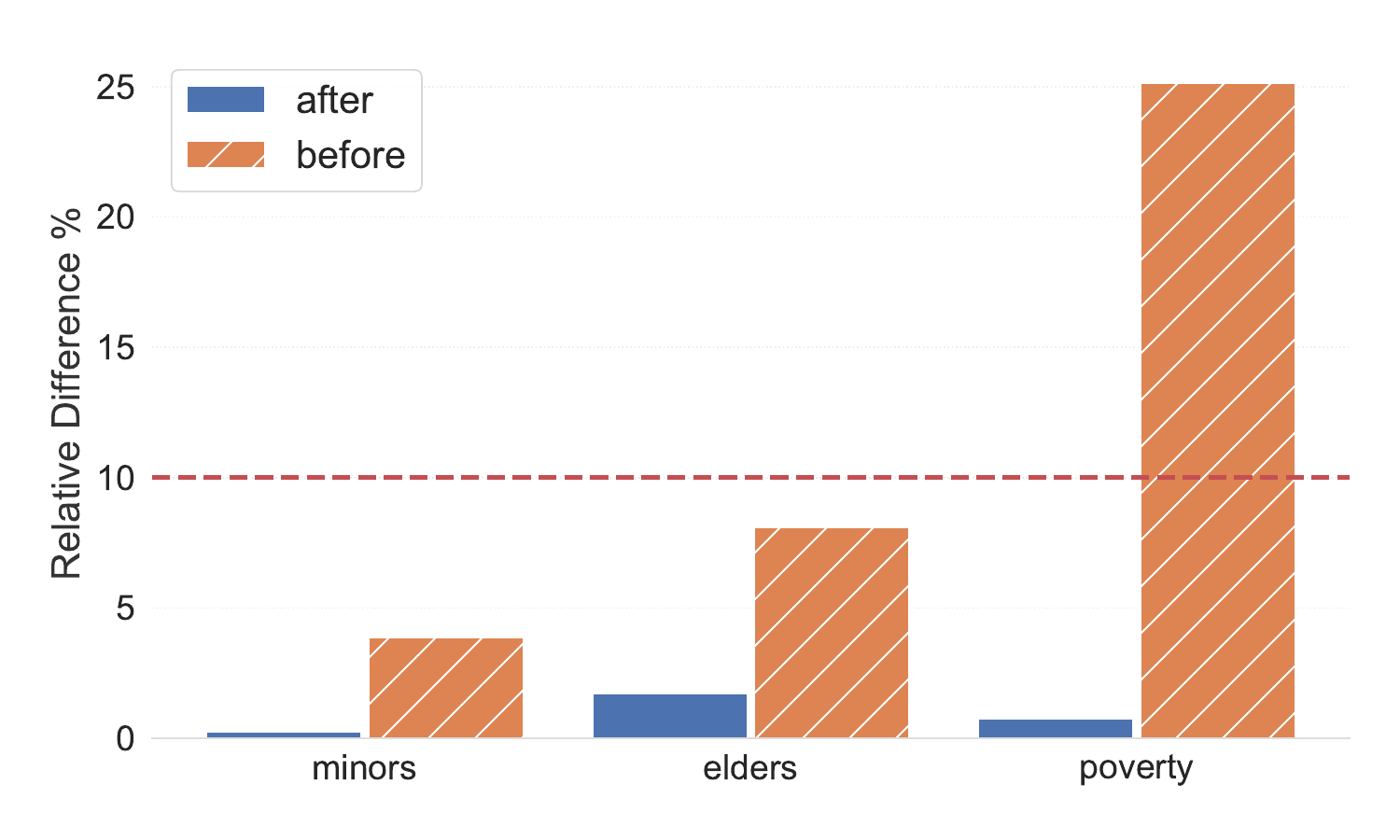}}
      \end{subfigure}
      \begin{subfigure}[Confounding factors of male rate.]{
        \includegraphics[height=0.18\textwidth]{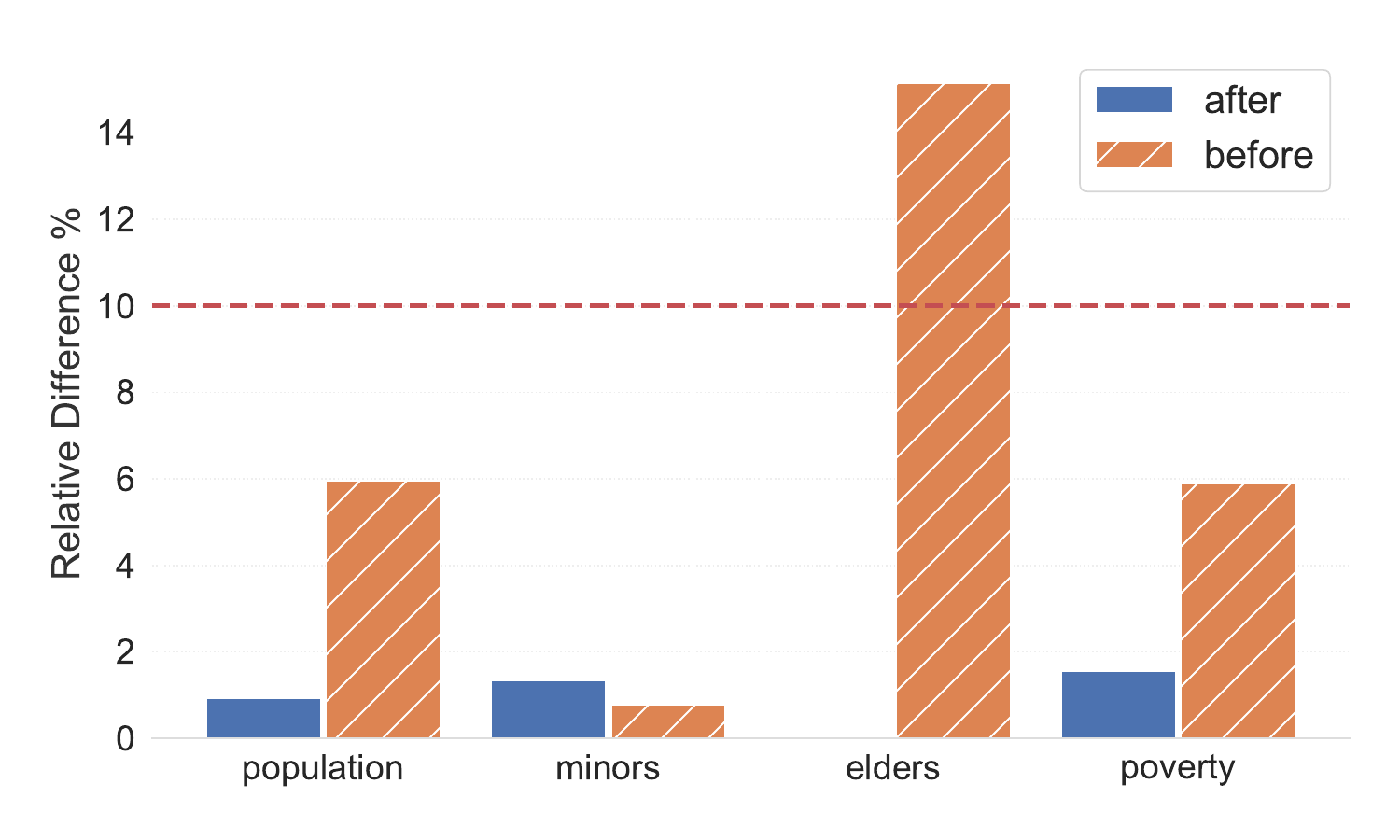}}
      \end{subfigure}
      \begin{subfigure}[Confounding factors of female rate.]{
        \includegraphics[height=0.18\textwidth]{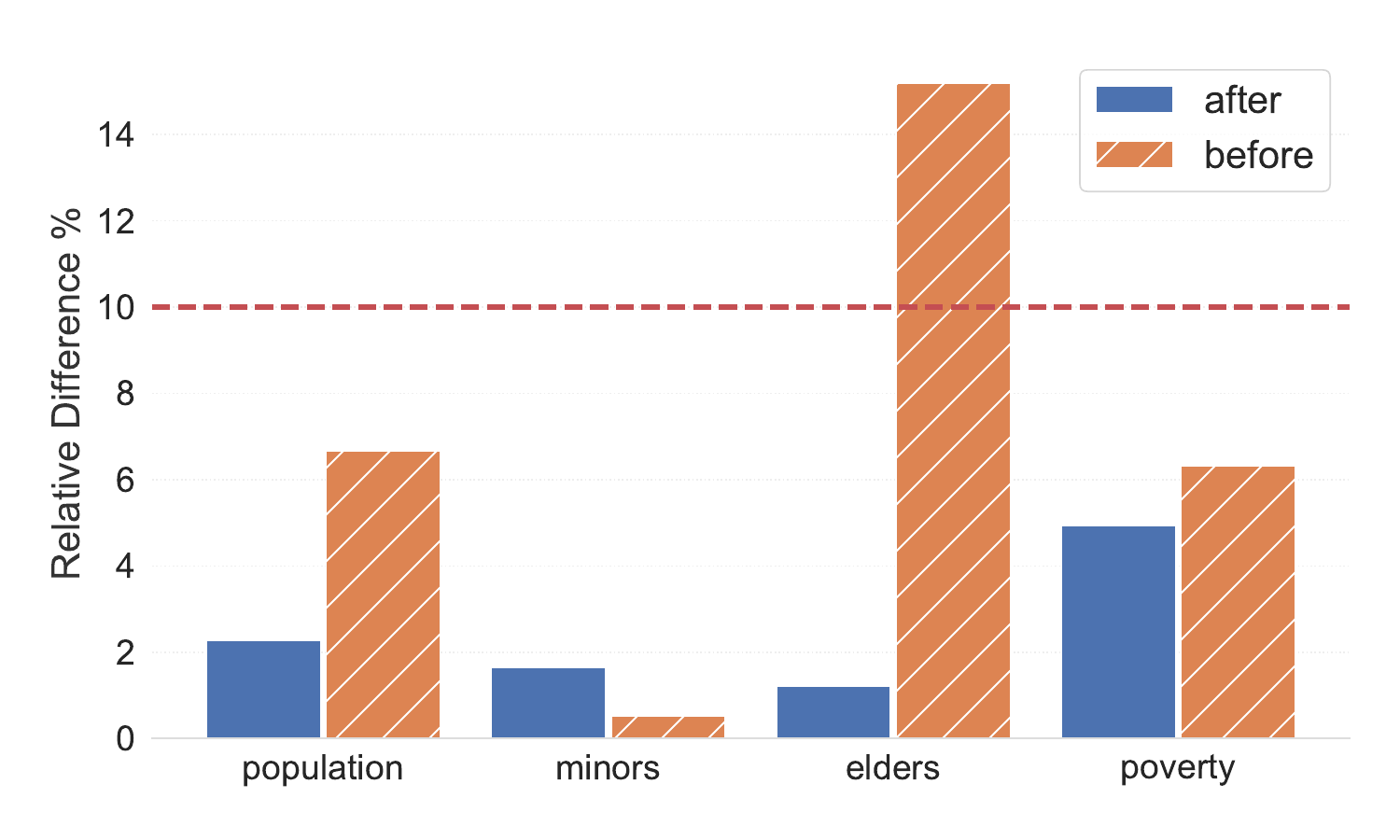}}
      \end{subfigure}
      \begin{subfigure}[Confounding factors of elders rate.]{
        \includegraphics[height=0.18\textwidth]{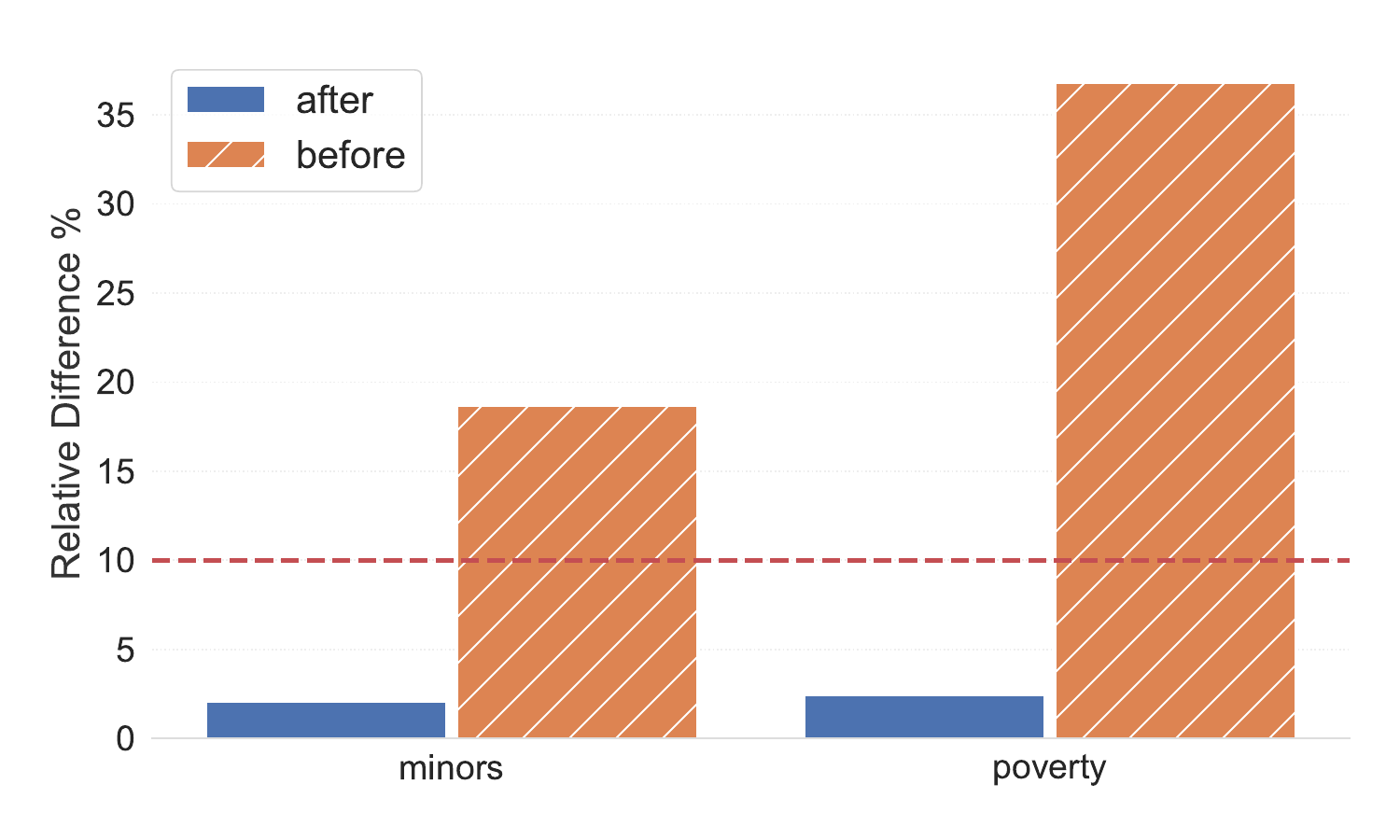}}
      \end{subfigure}
      \begin{subfigure}[Confounding factors of median age.]{
        \includegraphics[height=0.18\textwidth]{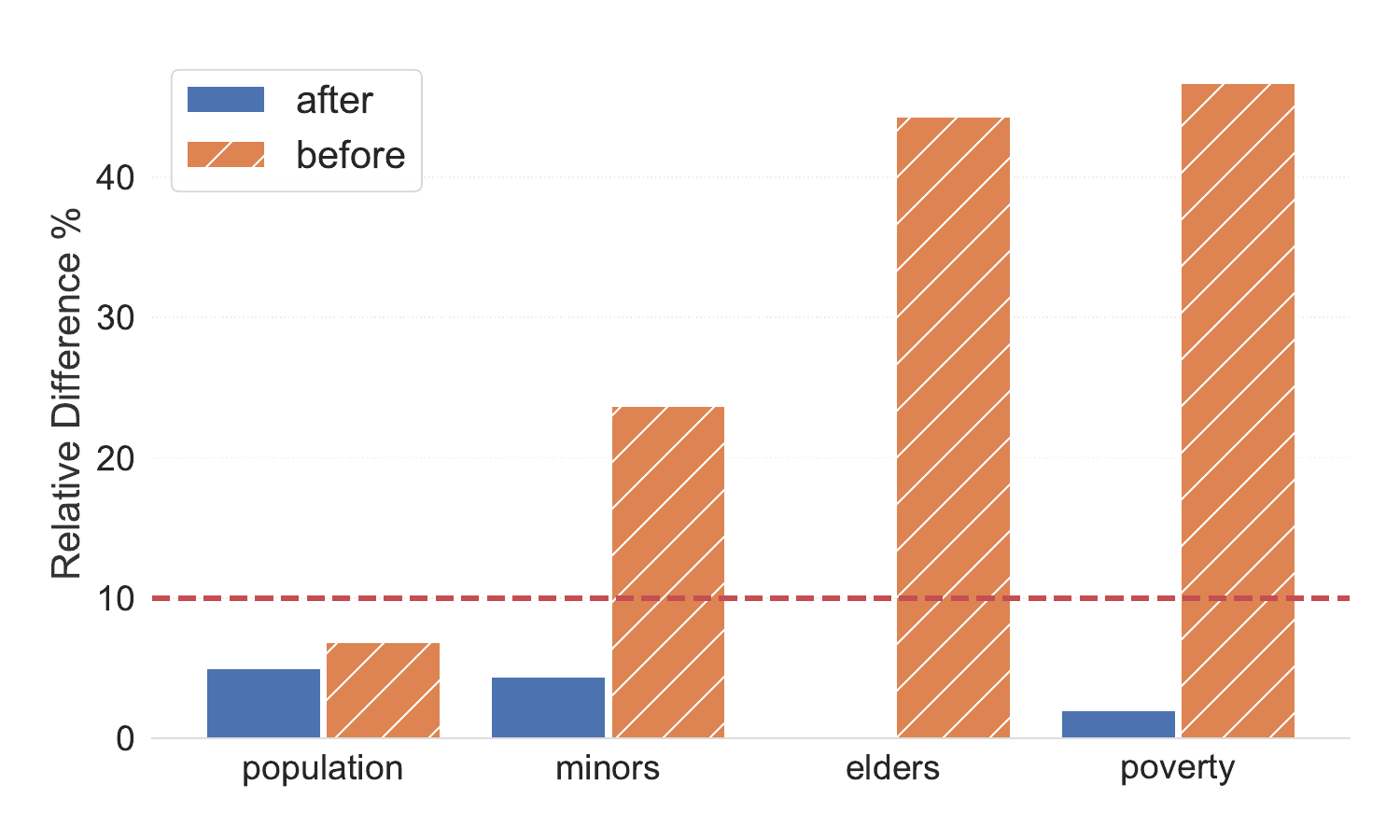}}
      \end{subfigure}
      \begin{subfigure}[Confounding factors of transport locations.]{
        \includegraphics[height=0.18\textwidth]{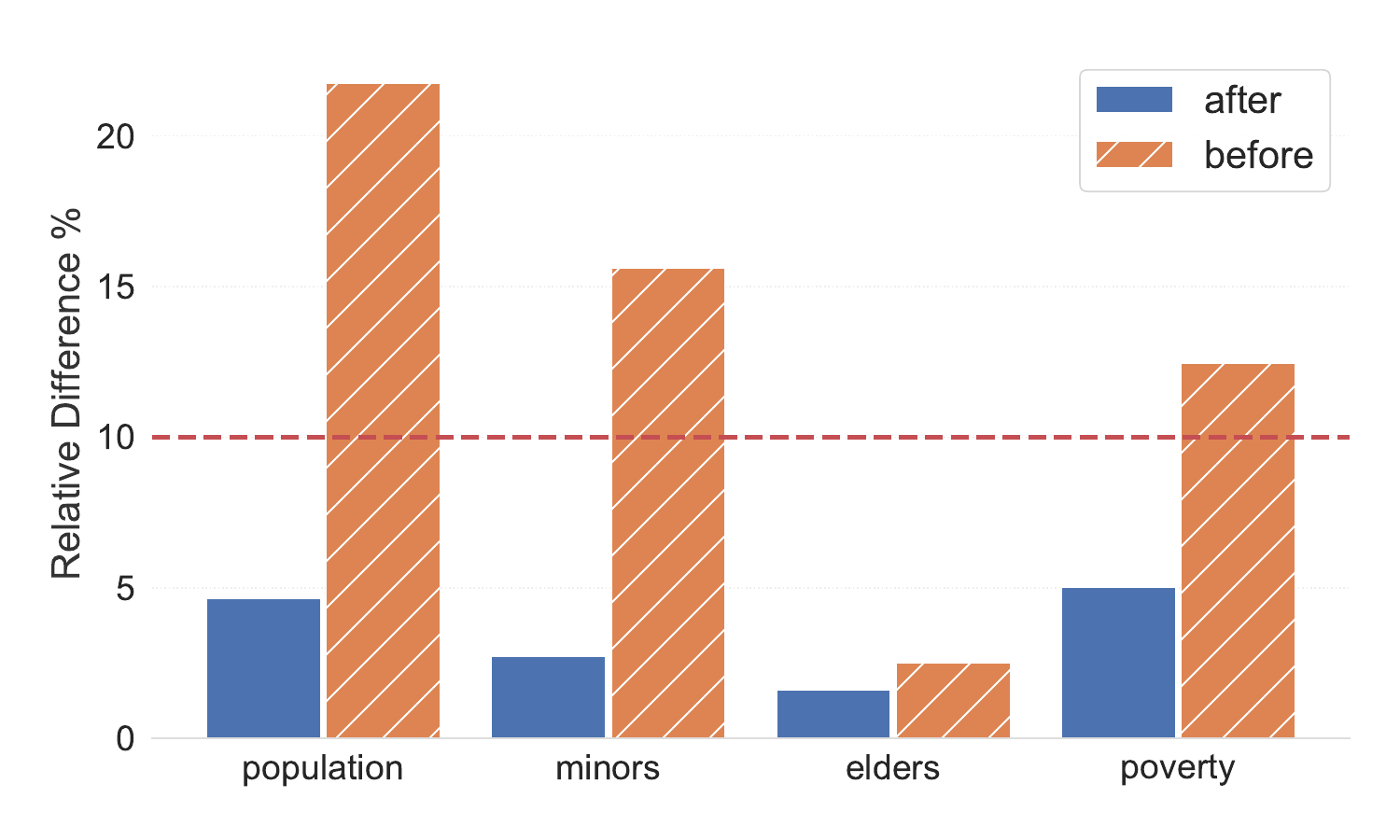}}
      \end{subfigure}
      \begin{subfigure}[Confounding factors of entertainment locations.]{
        \includegraphics[height=0.18\textwidth]{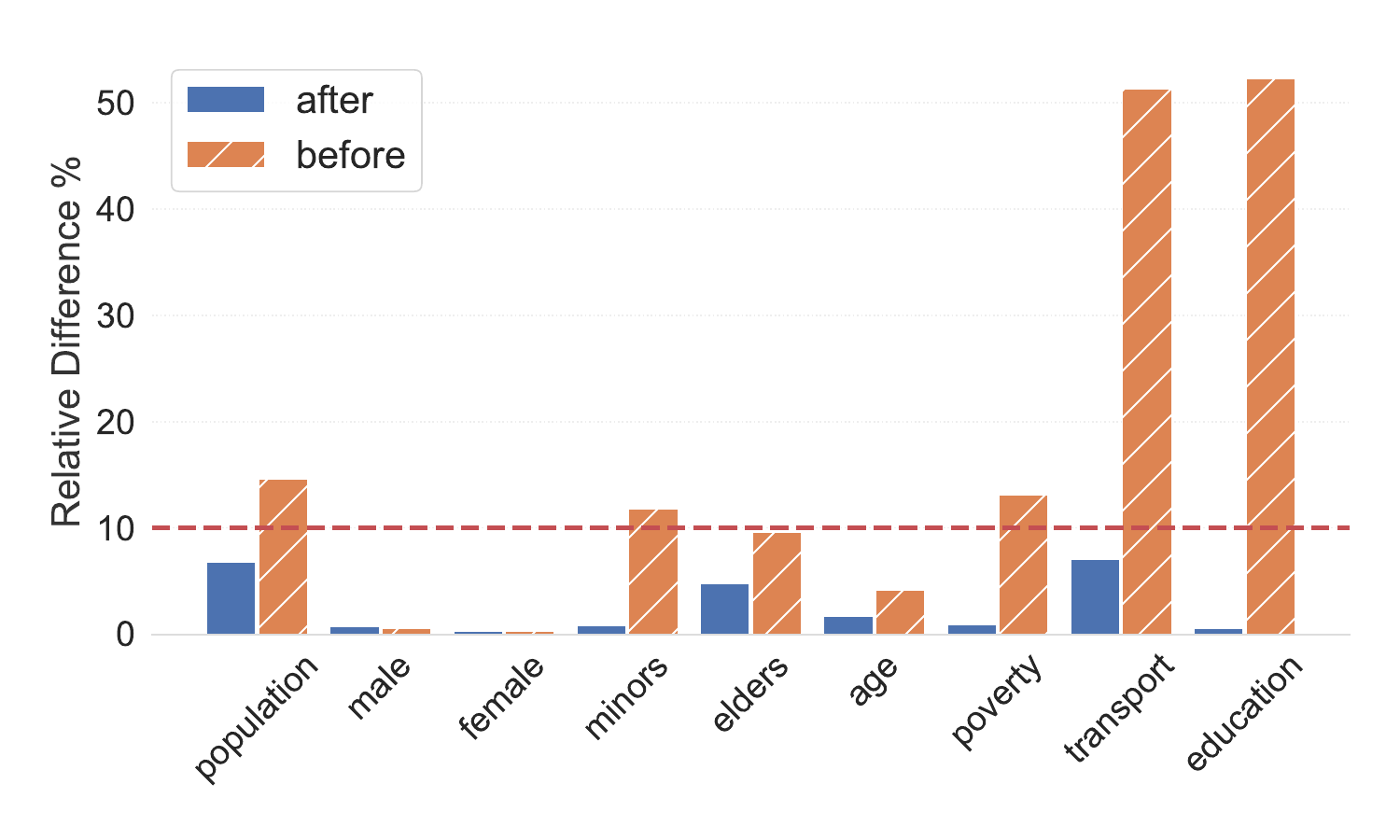}}
      \end{subfigure}
      \begin{subfigure}[Confounding factors of catering locations.]{
        \includegraphics[height=0.18\textwidth]{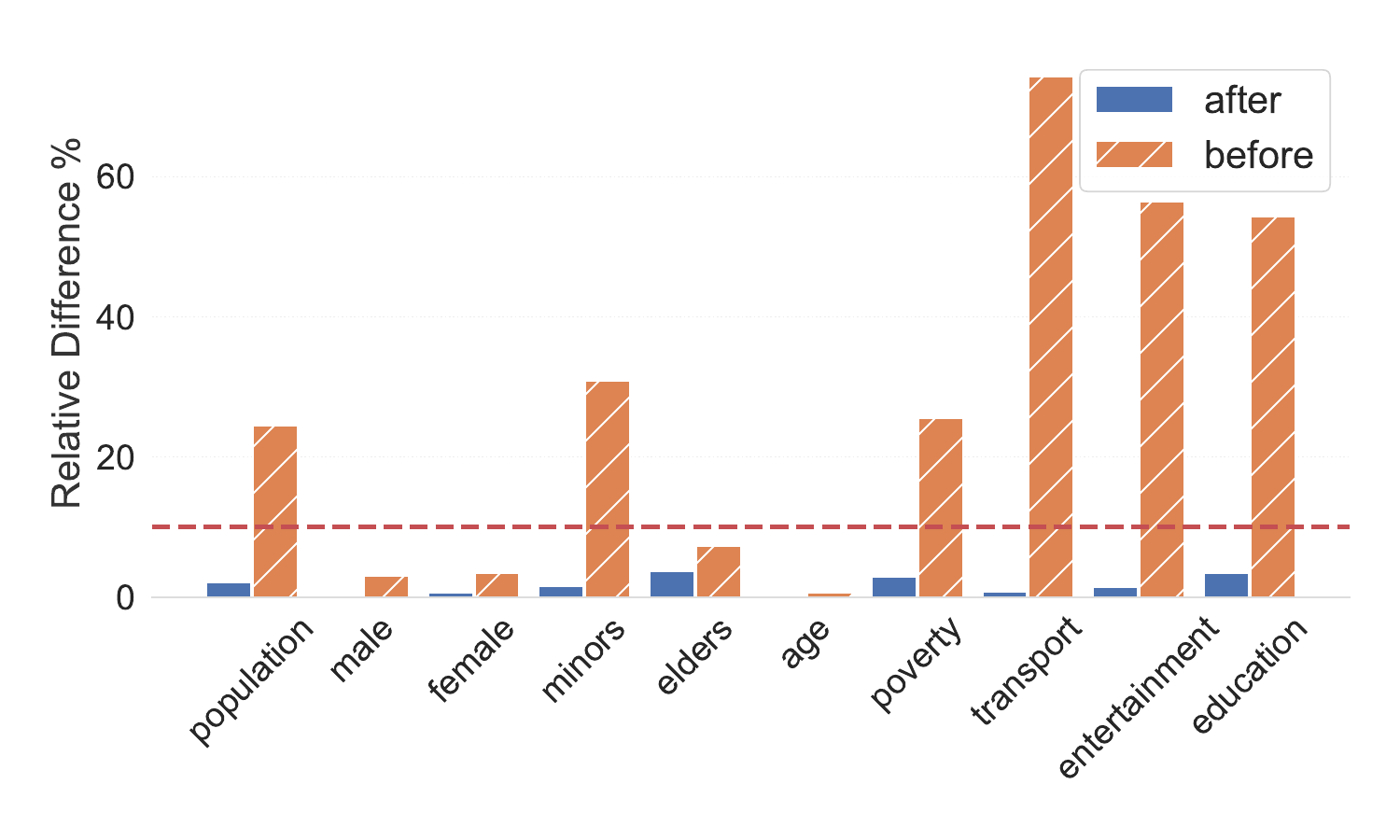}}
      \end{subfigure}
      \begin{subfigure}[Confounding factors of education locations.]{
        \includegraphics[height=0.18\textwidth]{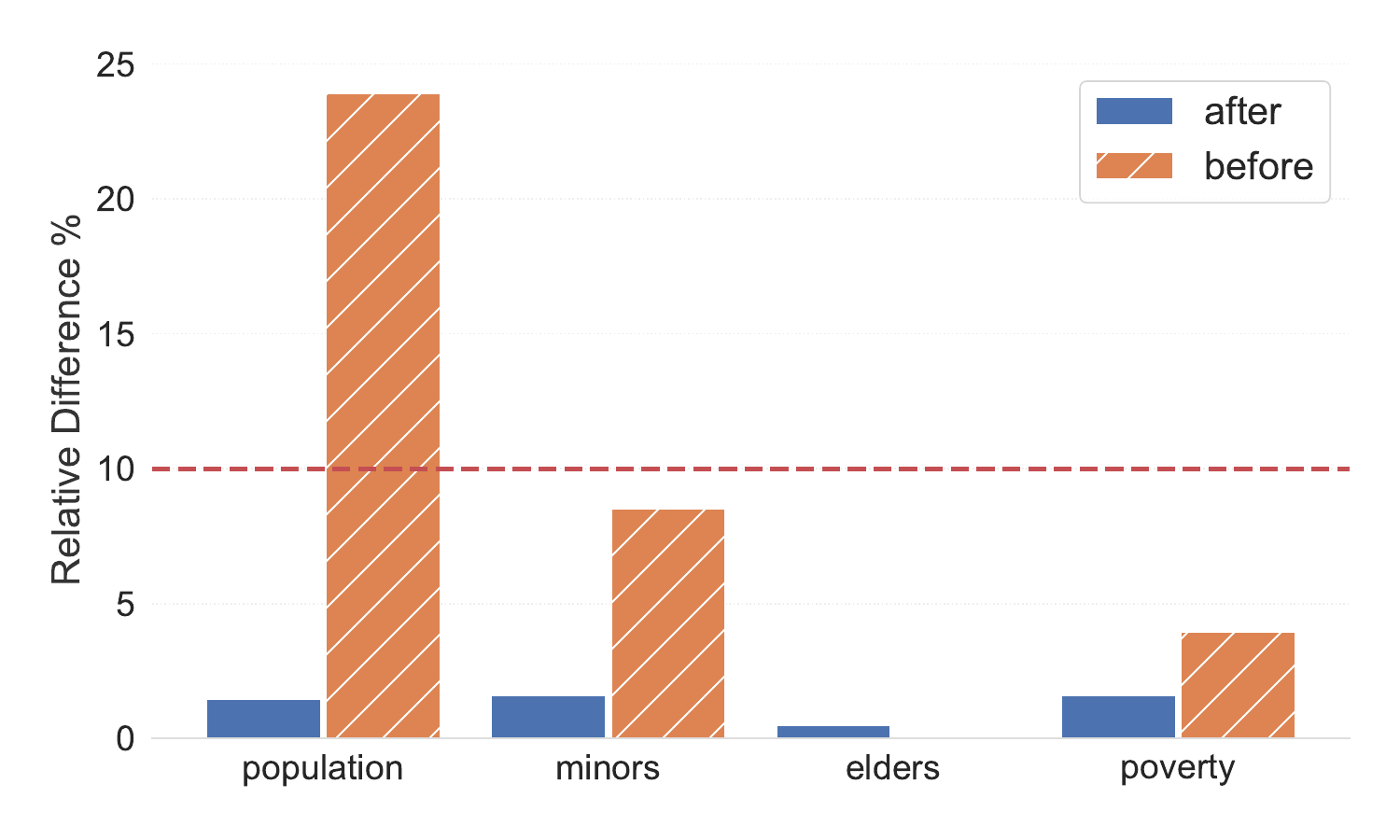}}
      \end{subfigure}
      \begin{subfigure}[Confounding factors of service locations.]{
        \includegraphics[height=0.22\textwidth]{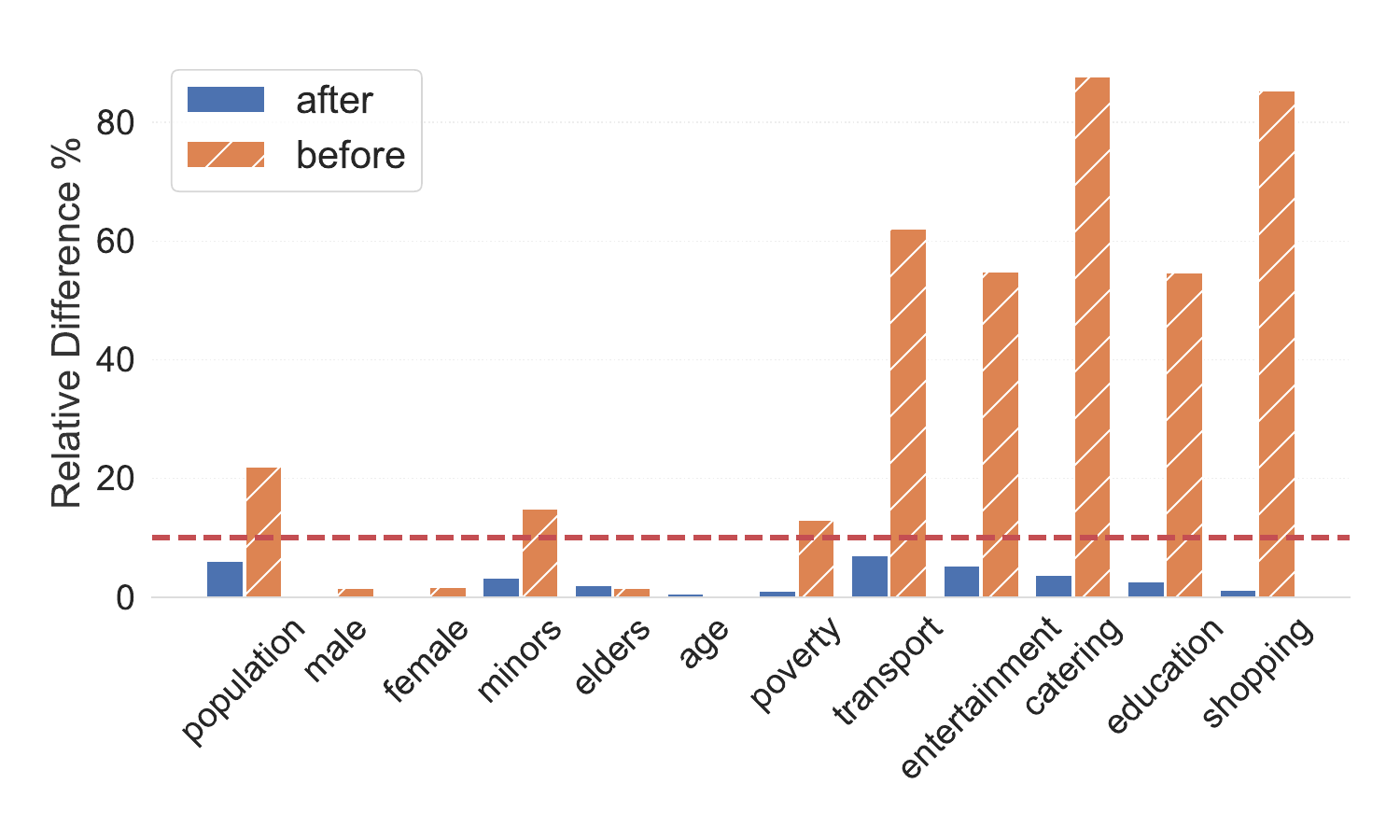}}
      \end{subfigure}
      \begin{subfigure}[Confounding factors of shopping locations.]{
        \includegraphics[height=0.22\textwidth]{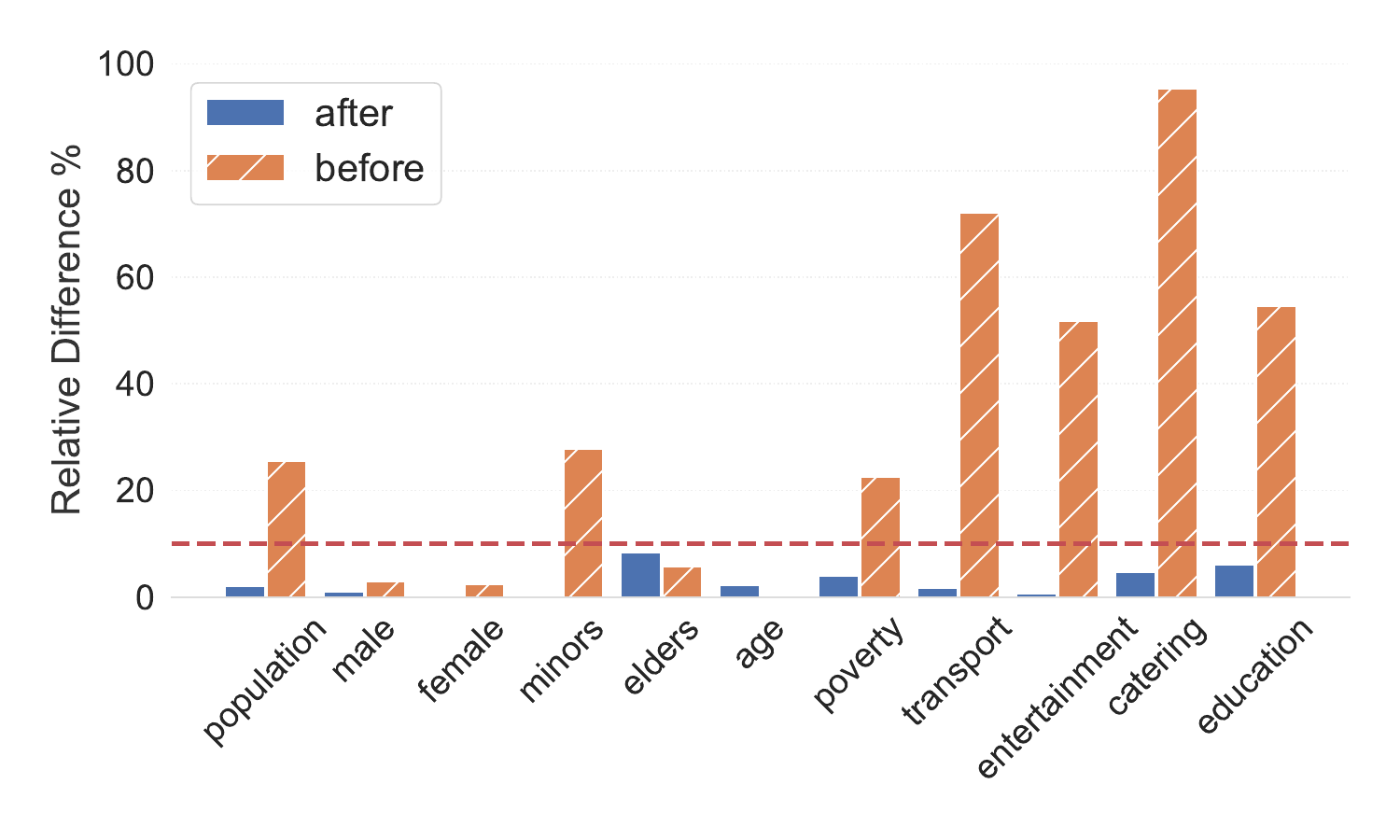}}
      \end{subfigure}
      \caption{Differences between confounding factors before and after applying propensity score matching. Blue bars are the relative difference between confounding factors on matched census tract pairs. Orange bars are the relative difference between confounding factors on census factors with top half and bottom half treatment levels. High relative difference implies large confounding effect.}
      \label{fig:balancing}
\end{figure*}

The adjacency matrix of the causal graph learned by the causal discovery algorithm is represented in Figure~\ref{fig:causal_graph_matrix}. The blue squares denote that the factor in the corresponding row is the cause of the factor in the corresponding column. For instance, minors rate is the cause of transportation and mobility, while transportation locations is the cause of mobility.
Therefore, minors rate is considered as one of the confounding factors in the causal edge from transportation locations to mobility.
Totally, there are 11 factors confounded by other factors.
The propensity score matching approach is leveraged to balance confounding factors.
First, we check the credibility of propensity score matching by comparing the balance on confounding factors before and after matching.
The balance before matching is measured by the difference between average values of confounding factors,
on regions with a treatment level over 2 (top half) and regions with a treatment level of less than 3 (lower half).
The balance after matching is measured by the average values of confounding factors,
on regions with a higher treatment level in each matched pair and regions with a lower treatment level in each pair. The relative difference is shown in Figure~\ref{fig:balancing}. 

Aside from the de-confounded causal effects, more insightful conclusions can be drawn from significant causal edges, especially the causal effects on urban mobility factors. The mobility flow in a region characterize its connectivity and interaction with other regions in the city, which is crucial for commercial, capital, and information communication. Attractive regions can have better economic gains.
In Figure~\ref{fig:significant_causal}, we can observe that {\em more transportation, catering, education, service, and shopping locations can increase the region's mobility flow, while more proportion of minors, elders, and poverty population and higher median age will decrease mobility flow}. These findings suggest that locations can attract more visitors, but local minors, elders, and people living in poverty cannot contribute to mobility flow in the urban space. As for the proportion of people taking public transportation, {\em more population, higher poverty ratio, lower minors and elders rate, and less education locations ratio will increase public transportation rate}, implying that poorer, middle-aged people are taking more public transportation. Finally, {\em higher minors and elders ratio, higher poverty ratio, and less transportation locations will lead to a higher travel time}. This may indicate that poorer people need to travel to further destinations to work, and transportation locations can improve travel speed.

We can observe from Figure~\ref{fig:balancing} that the relative difference of confounding factors on matched pairs are lower than 10\%, while the original difference can be greater than 90\%. The great improvements confirm the credibility of the propensity score matching procedure. Therefore, we can estimate the ATE by averaging individual treatment effects, which is the difference in outcome divided by the difference in treatment level in each matched pair. We also test the significance of each causal effect by two-sided t-test on the average of individual treatment effects. Edges with significance levels \emph{p < 0.05} are considered significant. In Figure~\ref{fig:significant_causal}, we draw the direction of each significant causal edge in the causal graph. For un-confounded edges like minors rate affecting the population, the direction of its causal effect is considered the same as the direction of correlation coefficients. For confounded edges, only ones with significant causal effects are kept in the graph.
Consequently, blue squares denote positive effects, and red squares represent negative effects.

Compelling findings can be drawn from the directions of causal effects. First, {\em some of the causal effects have different directions with correlation coefficients}. By comparing the colors in Figure~\ref{fig:significant_causal} and Figure~\ref{gra:corr}, we can summarize the following causal relations that are different from observed correlations:
\begin{itemize}
    \item elders rate has a positive causal effect on population, 
    \item elders rate has a negative causal effect on education locations, 
    \item median age has a negative causal effect on population mobility,
    \item median age has a negative causal effect on service locations, 
    \item median age has a negative causal effect on shopping locations,
    \item education locations has a negative causal effect on the proportion of people traveling by public transport. 
\end{itemize}

\begin{figure*}[]
  \centering
      \begin{subfigure}[]{
      \label{fig:cf1}
        \includegraphics[height=0.2\textwidth]{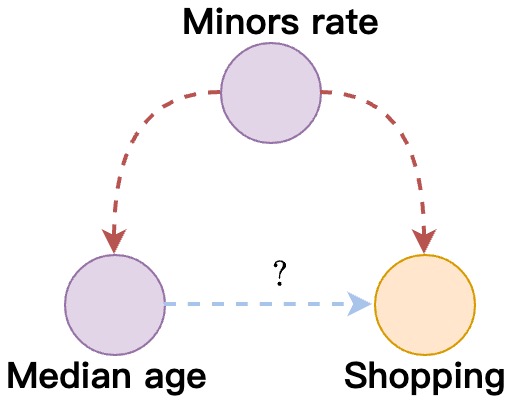}}
      \end{subfigure}
      \begin{subfigure}[]{
      \label{fig:cf2}
        \includegraphics[height=0.2\textwidth]{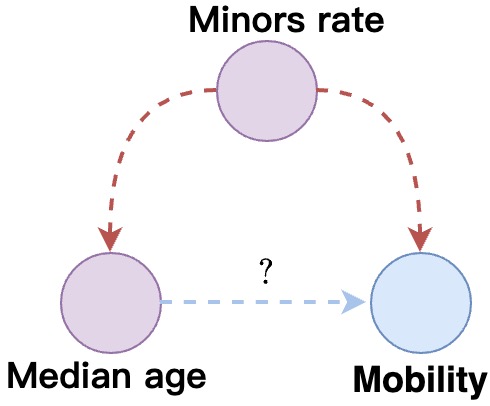}}
      \end{subfigure}
      \begin{subfigure}[]{
      \label{fig:cf3}
        \includegraphics[height=0.2\textwidth]{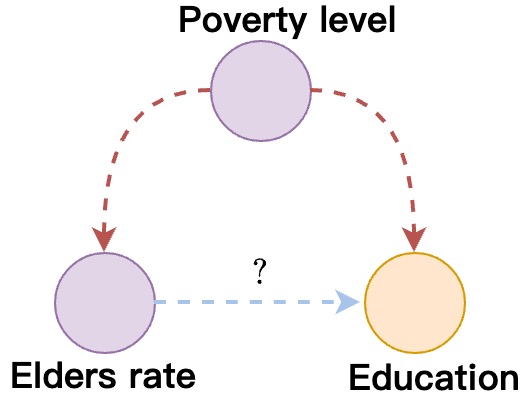}}
      \end{subfigure}
      \caption{Examples of confounded causal effects. }
      \label{fig:cf}
\end{figure*}

By contrast, correlation analysis provides opposite results. In Figure~\ref{fig:cf}, we present three examples of confounding factors influencing correlation analysis and causal effects.
As shown in Figure~\ref{fig:cf1}, {\em urban regions with higher median age will cause fewer shopping locations located in that region}. However, the minors rate in the region negatively affect both median age and the amount of shopping locations, creating a confounded positive correlation between median age and shopping locations. In Figure~\ref{fig:cf2}, {\em urban regions with higher median age will cause less population mobility}. Note that minors rate simultaneously affect both median age and mobility negatively, which confounds them be positively correlated. In Figure~\ref{fig:cf3}, {\em urban regions with higher elders rate cause fewer education locations}. This effect is confounded by the under poverty level rate, which negatively affects elders rate and education locations. 

\subsection{Performance of Urban Causal Prediction}

Identified urban causal relations can characterize factors' contributions to the production of other factors. This procedure implies strong connections between factors, which cannot be presented by simple correlation. We leverage the significance level of estimated causal effects for the input selection of urban prediction models. Specifically, we predict three mobility factors, \emph{i.e.}, population mobility, public transportation rate, and mean travel time to work. We consider two widely-used prediction models in supervised learning: linear regression and multi-layer perceptron (MLP). To confirm the robustness of our input selection strategy under the scenario of different training and test distribution, we first conduct five small-sample experiments,
with training sets only consisting of 20\% of census tracts.
We use the Rooted Mean Square Error (RMSE) and mean absolute error (MAE) to measure the performance on test sets.
The widely-used input selection strategies for supervised learning to prevent overfitting include L1 regularization~\cite{tibshirani1996regression} and univariate statistical selection.
We compare the causal significance based input selection with the following combinations of inputs:
1) all factors except for mobility behaviors (L1-regularized), 2) factors with a significant correlation (\emph{p<0.05}) with the predicted factor, and 3) ancestor nodes of the mobility behavior factor in the causal graph before pruning insignificant edges.

The experimental results are shown in Table~\ref{tab:prediction}.
Here, `All factors' corresponds to predicting mobility factors with all other factors and L1 regularization. `Correlation based' represents selecting factors with a significant correlation coefficient (\emph{p<0.05}) with the mobility factor as input. `Causal ancestor' is the combination of factors that locates on causal path targeting the mobility factor as shown in Figure~\ref{fig:causal_graph_matrix}.
`Causal significance' is the proposed selection strategy. Each column is a metric of a specific prediction method (linear regression or MLP) on four input combinations. 

\begin{table*}[]
\centering
\caption{Performance of models predicting mobility.}
\resizebox{0.99\textwidth}{!}
{\begin{tabular}{c|cccc|cccc|cccc}
\hline
\textbf{Predicted Factor}& \multicolumn{4}{c|}{\textbf{Mobility}}& \multicolumn{4}{c|}{\textbf{Public Transportation \%}}& \multicolumn{4}{c}{\textbf{Mean Travel Time}}\\ \hline
\multirow{2}{*}{\textbf{Inputs}} & \multicolumn{2}{c}{\textbf{MLP}} & \multicolumn{2}{c|}{\textbf{Linear Regression}} & \multicolumn{2}{c}{\textbf{MLP}} & \multicolumn{2}{c|}{\textbf{Linear Regression}} & \multicolumn{2}{c}{\textbf{MLP}} & \multicolumn{2}{c}{\textbf{Linear Regression}} \\
& RMSE& MAE& RMSE& MAE & RMSE  & MAE  & RMSE& MAE& RMSE & MAE & RMSE& MAE  \\ \hline
All factors & 3219.88 & 1262.45& 8543.71& 1960.98& 18.37&14.56& 15.33& 11.84& 8.41& 6.22& \textbf{7.05} & \textbf{4.94}\\
Correlation based & 3056.04& 1214.92& 4471.84&1656.30&16.82& 13.25& 21.15& 11.65& 8.19& 6.29& 12.95 & 5.16\\
Causal ancestor & 3104.74& 1248.57& 4627.26& 1694.36& 29.51&24.88& \textbf{14.15}& \textbf{11.09} & 22.55& 19.69& 7.73& 5.37\\
Causal significance & \textbf{3020.55}& \textbf{1165.74}& \textbf{3238.58}& \textbf{1534.09}& \textbf{15.87}& \textbf{12.53}& 14.17& 11.10& \textbf{7.34}& \textbf{5.26}& 7.73& 5.36\\
\hline
\end{tabular}
}
\label{tab:prediction}
\end{table*}

As for the prediction of population mobility, the proposed causal significance-based selection method achieves the best performance on both predictors and both metrics, while correlation-based selection method is the second best.
Moreover, when predicting the ratio of the population taking public transportation and mean travel time, our strategy outperforms baseline methods on the MLP regression predictor. Using causal ancestors in the original causal graph as input outperforms the causal significance selection method.
In addition, L1 regularization achieves the best performance when estimating mean travel time by linear regression.
From the above analysis, we can observe that {\em the causal significance-based input selection can achieve comparable performance with other powerful input selection approaches when predicting urban mobility factors with small training sizes}. In conclusion, causal significance-based input selection is a promising solution for further prediction tasks in urban computing, especially when factors are entangled and training samples are limited.

\begin{figure*}[]
  \centering
      \begin{subfigure}[RMSE on test set (test) and validation set (dev) of each epoch with significant cause factors of mobility as input.]{
        \includegraphics[width=0.48\textwidth]{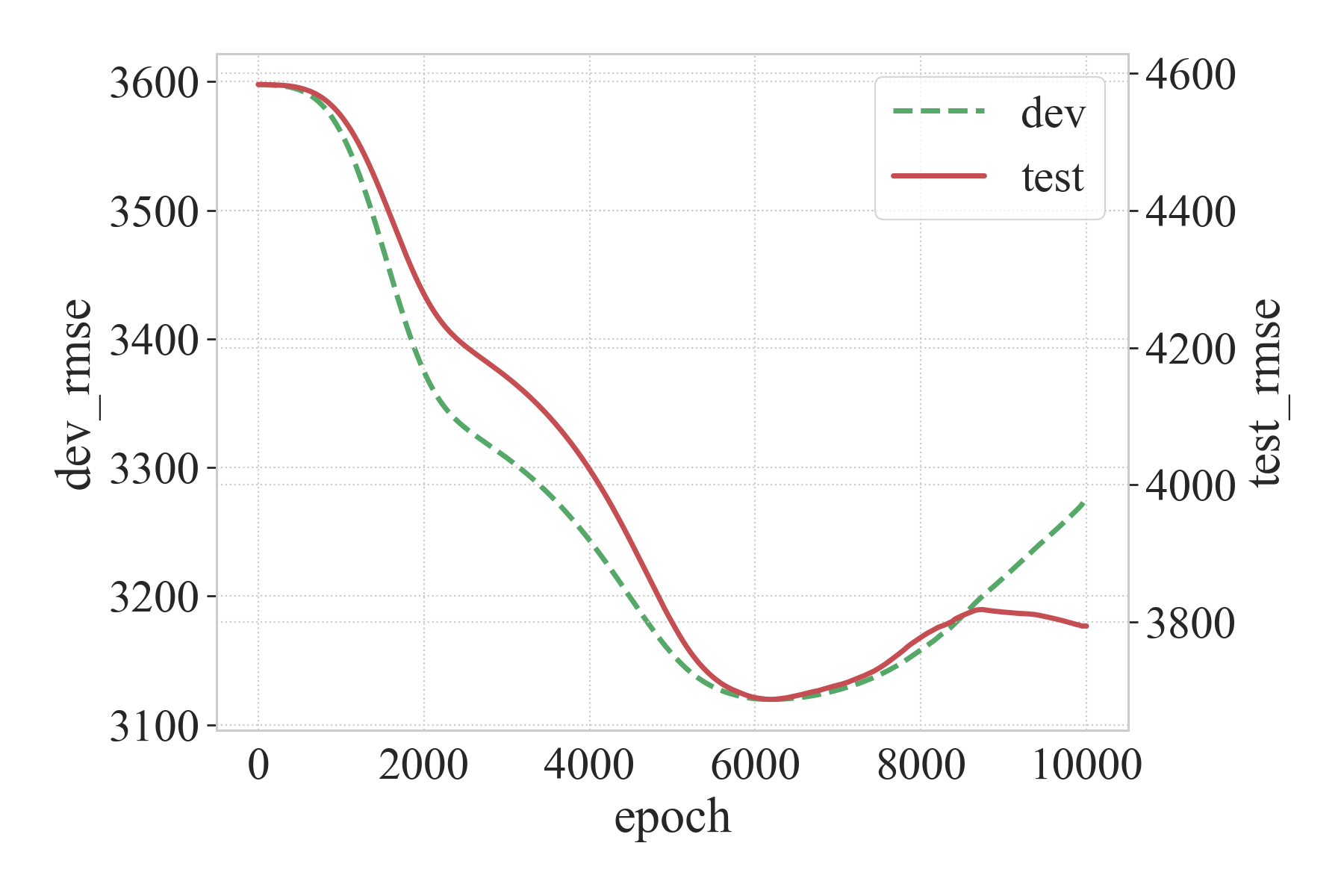}}
      \end{subfigure}
      \begin{subfigure}[RMSE on test set (test) and validation set (dev) of each epoch with significant correlated factors of mobility as input.]{
        \includegraphics[width=0.48\textwidth]{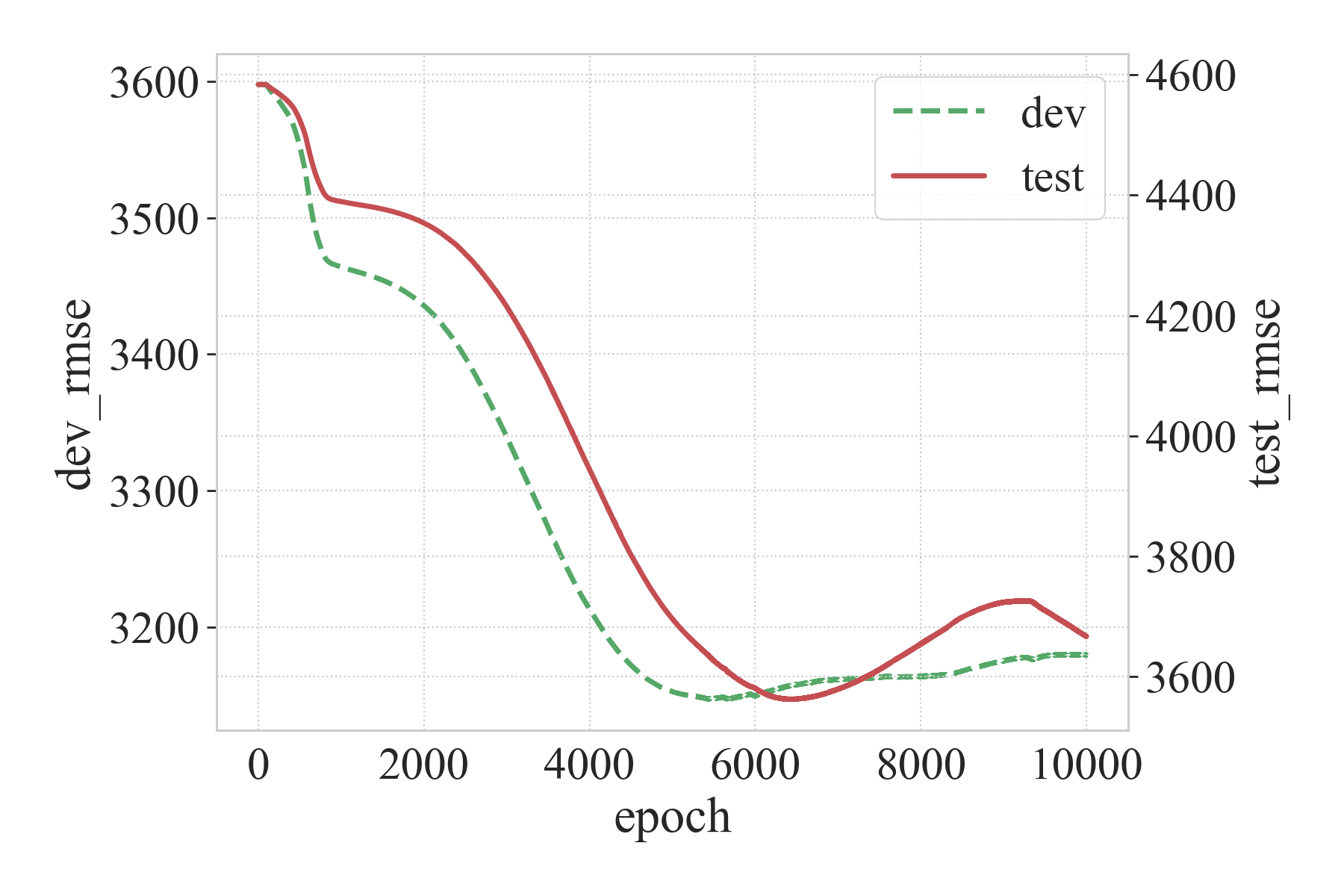}}
      \end{subfigure}
      \caption{Curves of RMSE of validation and test set in the prediction of population mobility with different input combinations.}
      \label{fig:overfit}
\end{figure*}

To test the strength of the causal significance based input selection for preventing overfitting, 
we plot the change of RMSE on the validation set and test set.
Here, we plot the change of RMSE on the validation set and test set in the training stage of MLP predictors,
with causal significant factors and correlation significant factors as inputs, respectively.
The proportions of the training set, validation set, and test set are 20\%, 20\%, and 60\%, respectively.
As shown in Figure~\ref{fig:overfit}, both strategies achieve the best performance at 6,000 epochs and overfit afterwards, meanwhile, the causal significance based method achieves more robust performance.

\begin{figure*}[]
  \centering
      \begin{subfigure}[RMSE on test sets.]{
      \label{fig:rmse}
        \includegraphics[width=0.48\textwidth]{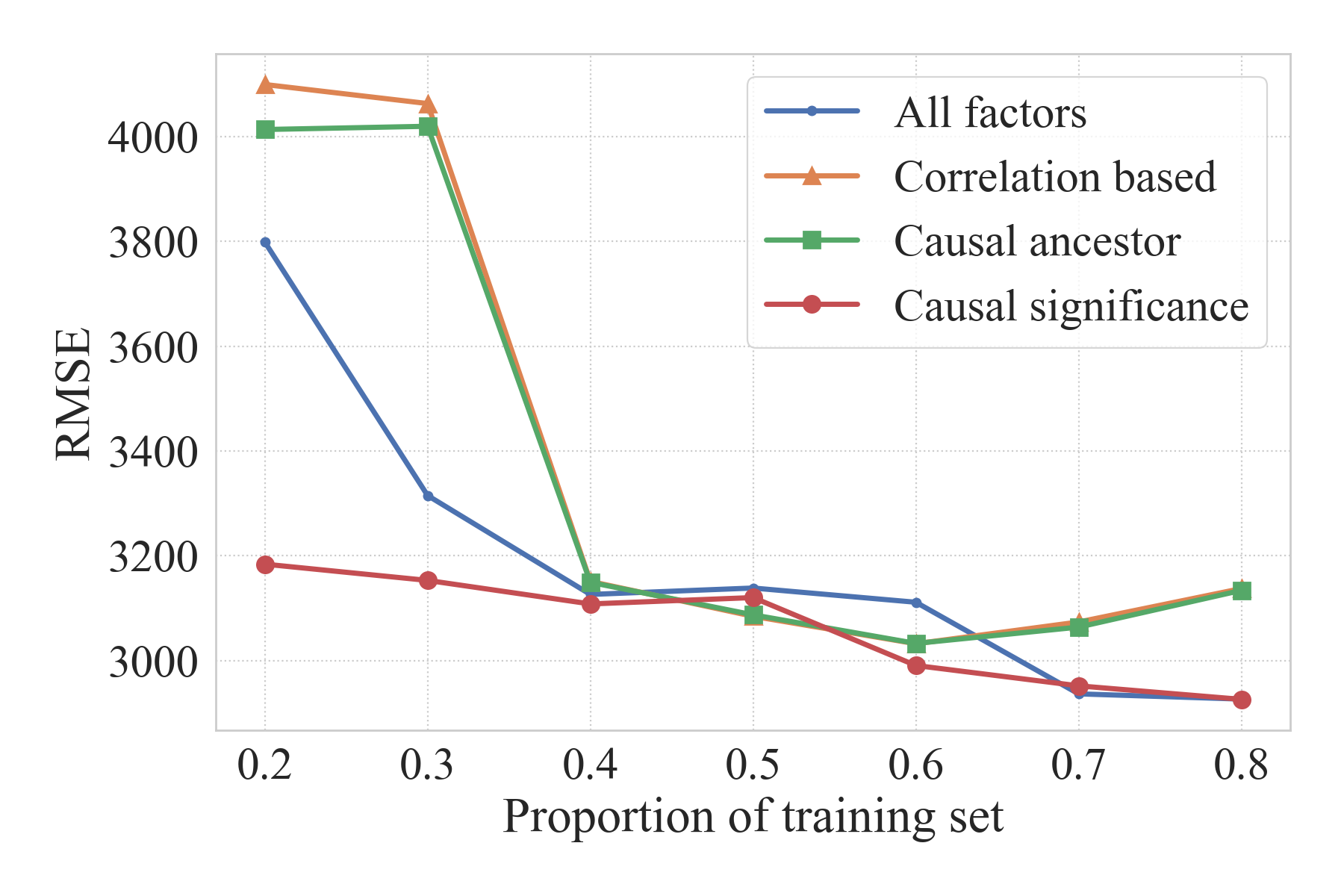}}
      \end{subfigure}
      \begin{subfigure}[MAE on training sets.]{
      \label{fig:mae}
        \includegraphics[width=0.48\textwidth]{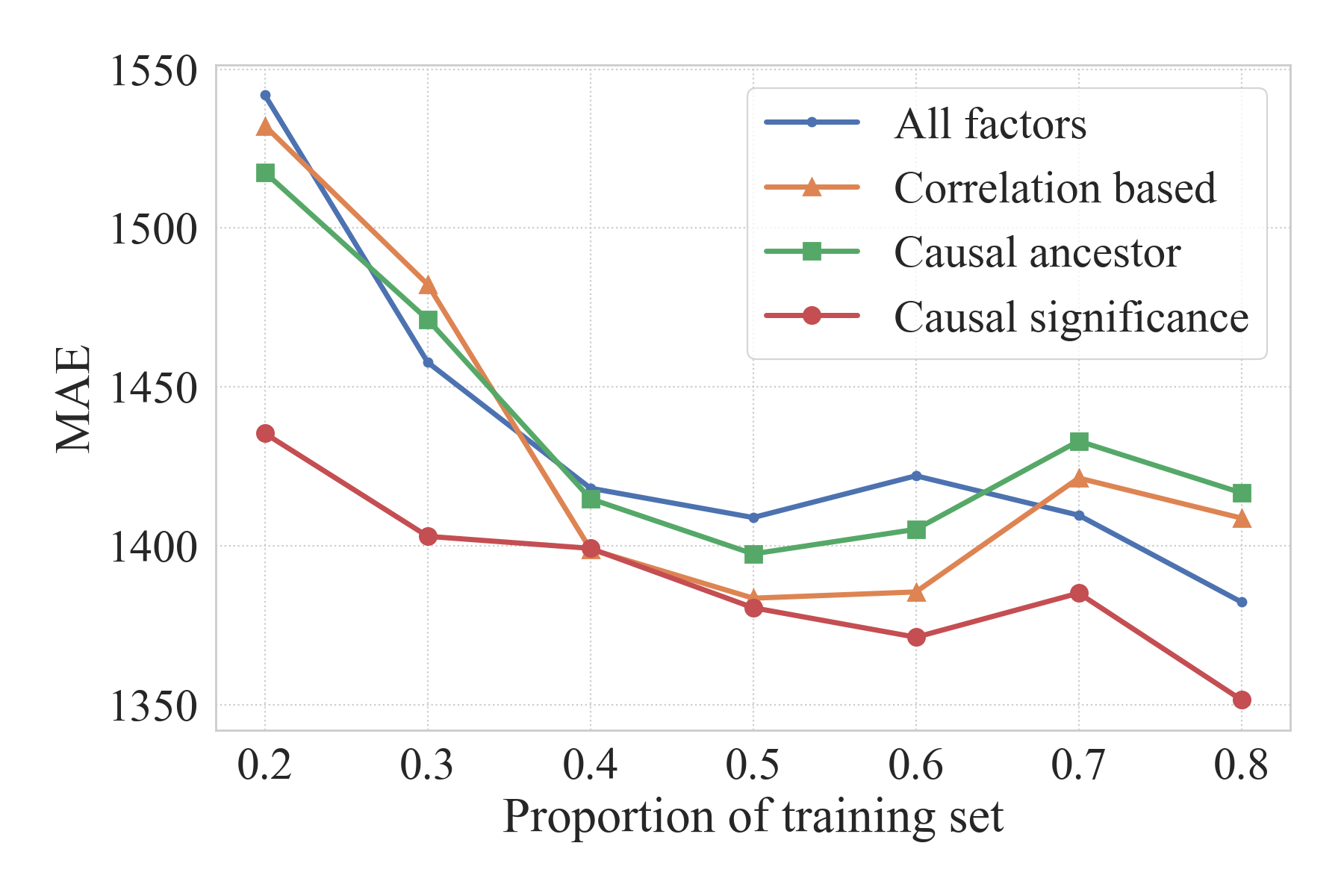}}
      \end{subfigure}
      \caption{Prediction performance under different training size.}
      \label{fig:pred}
\end{figure*}

Moreover, we analyze the performance of predicting population mobility with linear regression with four input combinations under different sizes of training samples. We conduct experiments with 7 different training sizes, ranging from 20\% to 80\%. In each experiment, we sample the training set with the corresponding training size, and fit a linear regression model on it.
Then, we measure the RMSE and MAE on the test set, repeating five times and taking the average value of metrics.
The curves of average RMSE and average MAE over the proportion of training set on four input combinations are presented in Figure~\ref{fig:pred}.

Here, all curves show a decline in RMSE or MAE when the training size increases.
We can observe from Figure~\ref{fig:rmse} that the average RMSE of the causal significance-based input selection outperforms other strategies, when the training set contains no more than half of all samples.
The average MAE of our method outperforms other strategies under all training sizes,
while its improvements gradually decreases when the training size is over 40\%.
The above observations suggest that the causal significance based input selection method can benefit urban mobility prediction with a small sampling set. To summarize, we use the significance of estimated causal effects in the proposed causal graph as a benchmark of pruning causal edges. It demonstrates that we can keep factors with significant causal effects on the mobility behavior factor to enhance the performance of predictions under small training size.

\section{Related Works and Discussion} \label{section:discussion}

\subsection{Relations between Factors in the Urban Space}
Since the concept of city has been evolving for thousands of years, urban space keeps gathering people, commerce, and knowledge, substantially transforming urban residents' lives, creating innumerable opportunities. According to the report from the United Nations Human Settlements Program, 56.2\% of the world population are urban populations in 2020~\cite{habitat2020world}. This ratio is astonishingly 82.6\% in Northern America. urban residents live, work, get educated, transport, and consume in cities, interacting with urban infrastructure and facilities. Therefore, it is important to shape the relationships between various factors within the urban space. 

Previous studies have elaborated on how factors in urban spaces are related~\cite{monsted2018phone, xu2020sume, liu2017point, yao2018exploiting,  xu2020deconstructing}. In~\cite{monsted2018phone}, the authors collected urban mobility behavior from cell phone data to extract basic personality traits. Xu \emph{et. al.}~\cite{xu2020sume} demonstrated that large-scale urban mobility behavior along with location information can be leveraged in inferring an individual's citizens' factors, including occupation, gender, age, education, and income. As for the inference of locations, Liu \emph{et. al.}~\cite{liu2017point} robustly modeled the demand of locations by integrating citizens' factors data, location profiles, and mobility data. Yao~\cite{yao2018exploiting} exploited mobility behavior data to empower location recommendation. Links between citizens' factors and mobility behaviors are studied widely in works on mobility inequality~\cite{gauvin2020gender,moro2021mobility, frias2012relation}. Gauvin \emph{et al.}~\cite{gauvin2020gender} discovered that women move in a smaller range and visit fewer locations than men in the urban space. Higher-income level is also correlated with higher mobility range~\cite{frias2012relation}. Moro ~\emph{et al.} discovered the strong correlation between income segregation and and urban mobility behaviours. The above studies successfully capture the correlation between citizens' factors, locations, and mobility behaviors.

In this paper, we attempt to leap over from correlation analysis to investigate a causal structure to depict the relations of factors under three basic urban elements: citizens, locations and urban mobility.
From the causal discovery algorithm via reinforcement learning, we propose a reasonable urban causal graph. The graph is three-tier in general, where citizens' factors affect neighborhood environment and they both are the causes of residents' mobility factors. Further analysis of causal effects provides more detailed information. Our findings consolidate previous theories and assumptions on the causal relations in the urban space. For example, locations can enhance local mobility levels, more residents living in poverty will cause higher travel time to work, less mobility flow, as well as fewer locations located. The above conclusions can serve as guidance to urban planning and regulation. Since causal relations imply that the `effect' will change when the `cause' varies, it is possible to enhance mobility level by intervening its parental node in the causal graph. At the census tract level, it can attract more mobility flow within the urban space by opening more locations, promoting its commercial and social values.

\subsection{Urban Causal Analysis}
The core of the causal analysis is to discover the causal relationship and causal effects between factors  from observational data. It  has been applied in many fields, such as health care~\cite{gershon2021informing,hu2021estimation,smith2021pluralistic,zhang2021quantifying}, recommendation systems~\cite{xu2021learning,zheng2021disentangling,xu2021causal}, urban computing~\cite{deng2021compass,baganz2021causal,myrovali2022spatio}, social networks~\cite{zheleva2021causal,sadri2021causal,emmert2021data}, \etc. Introducing causal analysis into urban computing has great potential. Conventional urban computing research often uses correlation analysis to measure the relationship between various factors in the city, which may leads to confusing and contrary conclusions due to confounding factors~\cite{lu2018inferring,shi2021attentional,yue2017measurements}. Moreover, methods based on correlation analysis also greatly reduces the interpretability of results,
thus restricting the ability of the practical application~\cite{vanderweele2013definition,mcnamee2003confounding,greenland2001confounding}.
Causal analysis can eliminate the deviation caused by the confounding factors, thereby helping us better understand the city and improve the performance of urban computing tasks by discovering the causal relationship and the causal effect among the factors in the city~\cite{zheleva2021causal,pearl2009causal,pearl2010causal,imbens2015causal}. 

\subsection{Applying Causal Discovery and Inference in Urban Computing}
Urban computing aims to solve problems through the data generated in the city~\cite{zheng2014urban}.
In the field of urban computing, mobility prediction has attracted people's attention and can help solve many problems in the city, such as urban planning and location selection.
Existing studies~\cite{lu2018inferring,shi2021attentional,yue2017measurements} select some factors related to mobility in cities to make predictions, such as population and distributions of location.
However, these methods are based on the correlation between factors and mobility,
which can easily lead to problems such as information redundancy and overfitting of predictions
In this study, motivated by the deficiency of the correlation-based model, we propose models based on causal discovery and inference to make robust mobility predictions.
We construct a causal graph from observed factors and exploit causal inference to obtain the causal effect of each factor on mobility, and factor selection is performed on this basis to better predict mobility.
Our method is generalizable to other areas of urban computing. That is, the relationship between factors is characterized by a causal graph and causal effects among the factors in the city, and this relationship can be used to improve downstream tasks and the interpretability of the model. 

\subsection{Limitations}
There are many factors in the city, but our dataset cannot cover all the factors in the city. In causal discovery and inference, these hidden factors may play roles as hidden confounders to influence the modeling of the causal graphs and causal effects. In addition,  more complex mobility prediction model is expected to verify the effectiveness of our method and its potential for improvement.
In future work, we will consider introducing new methods and other datasets to make up for these shortcomings.

\section{Conclusions}
In this paper, we have studied causal relations among citizens, locations, and mobility in the urban space.
We collected 16 urban factors within 2,167 census tracts, from the city of New York, during 2019 to 2020.
We have further proposed a causal analytic framework to explore causal relations among urban factors.
In particular, a deep reinforcement learning model has been developed to discover the causal graph of urban factors. The obtained urban causal graph has a three-tier hierarchical structure,
with citizens on the top, locations in the middle, and mobility at the bottom.
Based on the constructed causal graph, we have calculated causal effect of each edge in the causal graph with propensity score matching to alleviate confounding effects. 
The estimated causal effects have provided a new dimension for understanding relations among urban factors
and some causal effects even deviate significantly from common understandings in previous correlation-based studies.
Correspondingly, we have removed the urban factors that do not show strong causal effects on others in experiments.
Extensive experimental results have validated the effectiveness of the proposed causal discovery and inference method, especially when the size of training sample is limited.
The proposed method achieved significant performance improvements in predicting urban mobility flows,
as it leverages discovered causal relations among urban factors.
In conclusion, we are the first to explore comprehensive causal relations in urban space,
focusing on causal discovery and inference among urban factors in three dimensions.
The causal graph will serve as a new gate to enable urban planning and urban construction.

\bibliographystyle{ACM-Reference-Format}
\bibliography{refer}


\begin{thebibliography}{83}


\ifx \showCODEN    \undefined \def \showCODEN     #1{\unskip}     \fi
\ifx \showDOI      \undefined \def \showDOI       #1{#1}\fi
\ifx \showISBNx    \undefined \def \showISBNx     #1{\unskip}     \fi
\ifx \showISBNxiii \undefined \def \showISBNxiii  #1{\unskip}     \fi
\ifx \showISSN     \undefined \def \showISSN      #1{\unskip}     \fi
\ifx \showLCCN     \undefined \def \showLCCN      #1{\unskip}     \fi
\ifx \shownote     \undefined \def \shownote      #1{#1}          \fi
\ifx \showarticletitle \undefined \def \showarticletitle #1{#1}   \fi
\ifx \showURL      \undefined \def \showURL       {\relax}        \fi
\providecommand\bibfield[2]{#2}
\providecommand\bibinfo[2]{#2}
\providecommand\natexlab[1]{#1}
\providecommand\showeprint[2][]{arXiv:#2}

\bibitem[\protect\citeauthoryear{Baganz, Schrenk, K{\"o}rner, Baganz, Keesman,
  Goddek, Siscan, Baganz, Doernberg, Monsees, et~al\mbox{.}}{Baganz
  et~al\mbox{.}}{2021}]%
        {baganz2021causal}
\bibfield{author}{\bibinfo{person}{G{\"o}sta~FM Baganz},
  \bibinfo{person}{Manfred Schrenk}, \bibinfo{person}{Oliver K{\"o}rner},
  \bibinfo{person}{Daniela Baganz}, \bibinfo{person}{Karel~J Keesman},
  \bibinfo{person}{Simon Goddek}, \bibinfo{person}{Zorina Siscan},
  \bibinfo{person}{Elias Baganz}, \bibinfo{person}{Alexandra Doernberg},
  \bibinfo{person}{Hendrik Monsees}, {et~al\mbox{.}}}
  \bibinfo{year}{2021}\natexlab{}.
\newblock \showarticletitle{Causal relations of upscaled urban aquaponics and
  the Food-Water-Energy Nexus—A Berlin case study}.
\newblock \bibinfo{journal}{\emph{Water}} \bibinfo{volume}{13},
  \bibinfo{number}{15} (\bibinfo{year}{2021}), \bibinfo{pages}{2029}.
\newblock


\bibitem[\protect\citeauthoryear{Barrett, Clements, Foerster, and
  Lvovsky}{Barrett et~al\mbox{.}}{2020}]%
        {barrett2020exploratory}
\bibfield{author}{\bibinfo{person}{Thomas Barrett}, \bibinfo{person}{William
  Clements}, \bibinfo{person}{Jakob Foerster}, {and} \bibinfo{person}{Alex
  Lvovsky}.} \bibinfo{year}{2020}\natexlab{}.
\newblock \showarticletitle{Exploratory combinatorial optimization with
  reinforcement learning}. In \bibinfo{booktitle}{\emph{Proceedings of the AAAI
  Conference on Artificial Intelligence}}, Vol.~\bibinfo{volume}{34}.
  \bibinfo{pages}{3243--3250}.
\newblock


\bibitem[\protect\citeauthoryear{Bello, Pham, Le, Norouzi, and Bengio}{Bello
  et~al\mbox{.}}{2016}]%
        {bello2016neural}
\bibfield{author}{\bibinfo{person}{Irwan Bello}, \bibinfo{person}{Hieu Pham},
  \bibinfo{person}{Quoc~V Le}, \bibinfo{person}{Mohammad Norouzi}, {and}
  \bibinfo{person}{Samy Bengio}.} \bibinfo{year}{2016}\natexlab{}.
\newblock \showarticletitle{Neural combinatorial optimization with
  reinforcement learning}.
\newblock \bibinfo{journal}{\emph{arXiv preprint arXiv:1611.09940}}
  (\bibinfo{year}{2016}).
\newblock


\bibitem[\protect\citeauthoryear{Bettencourt and West}{Bettencourt and
  West}{2010}]%
        {bettencourt2010unified}
\bibfield{author}{\bibinfo{person}{Luis Bettencourt} {and}
  \bibinfo{person}{Geoffrey West}.} \bibinfo{year}{2010}\natexlab{}.
\newblock \showarticletitle{A unified theory of urban living}.
\newblock \bibinfo{journal}{\emph{Nature}} \bibinfo{volume}{467},
  \bibinfo{number}{7318} (\bibinfo{year}{2010}), \bibinfo{pages}{912--913}.
\newblock


\bibitem[\protect\citeauthoryear{Bettencourt}{Bettencourt}{2013}]%
        {bettencourt2013origins}
\bibfield{author}{\bibinfo{person}{Lu{\'\i}s~MA Bettencourt}.}
  \bibinfo{year}{2013}\natexlab{}.
\newblock \showarticletitle{The origins of scaling in cities}.
\newblock \bibinfo{journal}{\emph{science}} \bibinfo{volume}{340},
  \bibinfo{number}{6139} (\bibinfo{year}{2013}), \bibinfo{pages}{1438--1441}.
\newblock


\bibitem[\protect\citeauthoryear{Bettencourt}{Bettencourt}{2021}]%
        {bettencourt2021introduction}
\bibfield{author}{\bibinfo{person}{Luis~MA Bettencourt}.}
  \bibinfo{year}{2021}\natexlab{}.
\newblock \bibinfo{booktitle}{\emph{Introduction to urban science: evidence and
  theory of cities as complex systems}}.
\newblock \bibinfo{publisher}{MIT Press}.
\newblock


\bibitem[\protect\citeauthoryear{Burnham and Anderson}{Burnham and
  Anderson}{2004}]%
        {burnham2004multimodel}
\bibfield{author}{\bibinfo{person}{Kenneth~P Burnham} {and}
  \bibinfo{person}{David~R Anderson}.} \bibinfo{year}{2004}\natexlab{}.
\newblock \showarticletitle{Multimodel inference: understanding AIC and BIC in
  model selection}.
\newblock \bibinfo{journal}{\emph{Sociological methods \& research}}
  \bibinfo{volume}{33}, \bibinfo{number}{2} (\bibinfo{year}{2004}),
  \bibinfo{pages}{261--304}.
\newblock


\bibitem[\protect\citeauthoryear{Caliendo and Kopeinig}{Caliendo and
  Kopeinig}{2008}]%
        {caliendo2008some}
\bibfield{author}{\bibinfo{person}{Marco Caliendo} {and}
  \bibinfo{person}{Sabine Kopeinig}.} \bibinfo{year}{2008}\natexlab{}.
\newblock \showarticletitle{Some practical guidance for the implementation of
  propensity score matching}.
\newblock \bibinfo{journal}{\emph{Journal of economic surveys}}
  \bibinfo{volume}{22}, \bibinfo{number}{1} (\bibinfo{year}{2008}),
  \bibinfo{pages}{31--72}.
\newblock


\bibitem[\protect\citeauthoryear{Chen, Thaipisutikul, and Shih}{Chen
  et~al\mbox{.}}{2020}]%
        {chen2020learning}
\bibfield{author}{\bibinfo{person}{Yi-Cheng Chen}, \bibinfo{person}{Tipajin
  Thaipisutikul}, {and} \bibinfo{person}{Timothy~K Shih}.}
  \bibinfo{year}{2020}\natexlab{}.
\newblock \showarticletitle{A learning-based POI recommendation with
  spatiotemporal context awareness}.
\newblock \bibinfo{journal}{\emph{IEEE Transactions on Cybernetics}}
  (\bibinfo{year}{2020}).
\newblock


\bibitem[\protect\citeauthoryear{Cheng, Chen, De~Vos, Lai, and Witlox}{Cheng
  et~al\mbox{.}}{2019}]%
        {cheng2019applying}
\bibfield{author}{\bibinfo{person}{Long Cheng}, \bibinfo{person}{Xuewu Chen},
  \bibinfo{person}{Jonas De~Vos}, \bibinfo{person}{Xinjun Lai}, {and}
  \bibinfo{person}{Frank Witlox}.} \bibinfo{year}{2019}\natexlab{}.
\newblock \showarticletitle{Applying a random forest method approach to model
  travel mode choice behavior}.
\newblock \bibinfo{journal}{\emph{Travel behaviour and society}}
  \bibinfo{volume}{14} (\bibinfo{year}{2019}), \bibinfo{pages}{1--10}.
\newblock


\bibitem[\protect\citeauthoryear{Deng, Weng, Xie, Bao, Zheng, Xu, Chen, and
  Wu}{Deng et~al\mbox{.}}{2021}]%
        {deng2021compass}
\bibfield{author}{\bibinfo{person}{Zikun Deng}, \bibinfo{person}{Di Weng},
  \bibinfo{person}{Xiao Xie}, \bibinfo{person}{Jie Bao}, \bibinfo{person}{Yu
  Zheng}, \bibinfo{person}{Mingliang Xu}, \bibinfo{person}{Wei Chen}, {and}
  \bibinfo{person}{Yingcai Wu}.} \bibinfo{year}{2021}\natexlab{}.
\newblock \showarticletitle{Compass: Towards Better Causal Analysis of Urban
  Time Series}.
\newblock \bibinfo{journal}{\emph{IEEE Transactions on Visualization and
  Computer Graphics}} (\bibinfo{year}{2021}).
\newblock


\bibitem[\protect\citeauthoryear{Emmert-Streib and Dehmer}{Emmert-Streib and
  Dehmer}{2021}]%
        {emmert2021data}
\bibfield{author}{\bibinfo{person}{Frank Emmert-Streib} {and}
  \bibinfo{person}{Matthias Dehmer}.} \bibinfo{year}{2021}\natexlab{}.
\newblock \showarticletitle{Data-driven computational social network science:
  Predictive and inferential models for Web-enabled scientific discoveries}.
\newblock \bibinfo{journal}{\emph{Frontiers in big Data}}  \bibinfo{volume}{4}
  (\bibinfo{year}{2021}), \bibinfo{pages}{10}.
\newblock


\bibitem[\protect\citeauthoryear{Fang, Yang, Wang, Fu, Song, Zhang, and
  Zhang}{Fang et~al\mbox{.}}{2019}]%
        {fang2019mac}
\bibfield{author}{\bibinfo{person}{Zhihan Fang}, \bibinfo{person}{Yu Yang},
  \bibinfo{person}{Shuai Wang}, \bibinfo{person}{Boyang Fu},
  \bibinfo{person}{Zixing Song}, \bibinfo{person}{Fan Zhang}, {and}
  \bibinfo{person}{Desheng Zhang}.} \bibinfo{year}{2019}\natexlab{}.
\newblock \showarticletitle{Mac: Measuring the impacts of anomalies on travel
  time of multiple transportation systems}.
\newblock \bibinfo{journal}{\emph{Proceedings of the ACM on Interactive,
  Mobile, Wearable and Ubiquitous Technologies}} \bibinfo{volume}{3},
  \bibinfo{number}{2} (\bibinfo{year}{2019}), \bibinfo{pages}{1--24}.
\newblock


\bibitem[\protect\citeauthoryear{Frias-Martinez, Virseda-Jerez, and
  Frias-Martinez}{Frias-Martinez et~al\mbox{.}}{2012}]%
        {frias2012relation}
\bibfield{author}{\bibinfo{person}{Vanessa Frias-Martinez},
  \bibinfo{person}{Jesus Virseda-Jerez}, {and} \bibinfo{person}{Enrique
  Frias-Martinez}.} \bibinfo{year}{2012}\natexlab{}.
\newblock \showarticletitle{On the relation between socio-economic status and
  physical mobility}.
\newblock \bibinfo{journal}{\emph{Information Technology for Development}}
  \bibinfo{volume}{18}, \bibinfo{number}{2} (\bibinfo{year}{2012}),
  \bibinfo{pages}{91--106}.
\newblock


\bibitem[\protect\citeauthoryear{Gauvin, Tizzoni, Piaggesi, Young, Adler,
  Verhulst, Ferres, and Cattuto}{Gauvin et~al\mbox{.}}{2020}]%
        {gauvin2020gender}
\bibfield{author}{\bibinfo{person}{Laetitia Gauvin}, \bibinfo{person}{Michele
  Tizzoni}, \bibinfo{person}{Simone Piaggesi}, \bibinfo{person}{Andrew Young},
  \bibinfo{person}{Natalia Adler}, \bibinfo{person}{Stefaan Verhulst},
  \bibinfo{person}{Leo Ferres}, {and} \bibinfo{person}{Ciro Cattuto}.}
  \bibinfo{year}{2020}\natexlab{}.
\newblock \showarticletitle{Gender gaps in urban mobility}.
\newblock \bibinfo{journal}{\emph{Humanities and Social Sciences
  Communications}} \bibinfo{volume}{7}, \bibinfo{number}{1}
  (\bibinfo{year}{2020}), \bibinfo{pages}{1--13}.
\newblock


\bibitem[\protect\citeauthoryear{Gershon, Lindenauer, Wilson, Rose, Walkey,
  Sadatsafavi, Anstrom, Au, Bender, Brookhart, et~al\mbox{.}}{Gershon
  et~al\mbox{.}}{2021}]%
        {gershon2021informing}
\bibfield{author}{\bibinfo{person}{Andrea~S Gershon}, \bibinfo{person}{Peter~K
  Lindenauer}, \bibinfo{person}{Kevin~C Wilson}, \bibinfo{person}{Louise Rose},
  \bibinfo{person}{Allan~J Walkey}, \bibinfo{person}{Mohsen Sadatsafavi},
  \bibinfo{person}{Kevin~J Anstrom}, \bibinfo{person}{David~H Au},
  \bibinfo{person}{Bruce~G Bender}, \bibinfo{person}{M~Alan Brookhart},
  {et~al\mbox{.}}} \bibinfo{year}{2021}\natexlab{}.
\newblock \showarticletitle{Informing healthcare decisions with observational
  research assessing causal effect. An official American Thoracic Society
  Research Statement}.
\newblock \bibinfo{journal}{\emph{American journal of respiratory and critical
  care medicine}} \bibinfo{volume}{203}, \bibinfo{number}{1}
  (\bibinfo{year}{2021}), \bibinfo{pages}{14--23}.
\newblock


\bibitem[\protect\citeauthoryear{Greenland and Morgenstern}{Greenland and
  Morgenstern}{2001}]%
        {greenland2001confounding}
\bibfield{author}{\bibinfo{person}{Sander Greenland} {and} \bibinfo{person}{Hal
  Morgenstern}.} \bibinfo{year}{2001}\natexlab{}.
\newblock \showarticletitle{Confounding in health research}.
\newblock \bibinfo{journal}{\emph{Annual review of public health}}
  \bibinfo{volume}{22} (\bibinfo{year}{2001}), \bibinfo{pages}{189}.
\newblock


\bibitem[\protect\citeauthoryear{Habitat}{Habitat}{2020}]%
        {habitat2020world}
\bibfield{author}{\bibinfo{person}{UN Habitat}.}
  \bibinfo{year}{2020}\natexlab{}.
\newblock \bibinfo{title}{WORLD CITIES REPORT 2020: The value of sustainable
  urbanization}.
\newblock
\newblock


\bibitem[\protect\citeauthoryear{Hasthanasombat and Mascolo}{Hasthanasombat and
  Mascolo}{2019}]%
        {hasthanasombat2019understanding}
\bibfield{author}{\bibinfo{person}{Apinan Hasthanasombat} {and}
  \bibinfo{person}{Cecilia Mascolo}.} \bibinfo{year}{2019}\natexlab{}.
\newblock \showarticletitle{Understanding the effects of the neighbourhood
  built environment on public health with open data}. In
  \bibinfo{booktitle}{\emph{The World Wide Web Conference}}.
  \bibinfo{pages}{648--658}.
\newblock


\bibitem[\protect\citeauthoryear{Heckerman, Meek, and Cooper}{Heckerman
  et~al\mbox{.}}{1999}]%
        {heckerman1999bayesian}
\bibfield{author}{\bibinfo{person}{David Heckerman},
  \bibinfo{person}{Christopher Meek}, {and} \bibinfo{person}{Gregory Cooper}.}
  \bibinfo{year}{1999}\natexlab{}.
\newblock \showarticletitle{A Bayesian approach to causal discovery}.
\newblock \bibinfo{journal}{\emph{Computation, causation, and discovery}}
  \bibinfo{volume}{19} (\bibinfo{year}{1999}), \bibinfo{pages}{141--166}.
\newblock


\bibitem[\protect\citeauthoryear{Herter, dos Santos, and Pinto}{Herter
  et~al\mbox{.}}{2014}]%
        {herter2014man}
\bibfield{author}{\bibinfo{person}{M{\'a}rcia~Maurer Herter},
  \bibinfo{person}{Cristiane~Pizzutti dos Santos}, {and}
  \bibinfo{person}{Diego~Costa Pinto}.} \bibinfo{year}{2014}\natexlab{}.
\newblock \showarticletitle{“Man, I shop like a woman!” The effects of
  gender and emotions on consumer shopping behaviour outcomes}.
\newblock \bibinfo{journal}{\emph{International Journal of Retail \&
  Distribution Management}} (\bibinfo{year}{2014}).
\newblock


\bibitem[\protect\citeauthoryear{Hu and Gu}{Hu and Gu}{2021}]%
        {hu2021estimation}
\bibfield{author}{\bibinfo{person}{Liangyuan Hu} {and}
  \bibinfo{person}{Chenyang Gu}.} \bibinfo{year}{2021}\natexlab{}.
\newblock \showarticletitle{Estimation of causal effects of multiple treatments
  in healthcare database studies with rare outcomes}.
\newblock \bibinfo{journal}{\emph{Health Services and Outcomes Research
  Methodology}} (\bibinfo{year}{2021}), \bibinfo{pages}{1--22}.
\newblock


\bibitem[\protect\citeauthoryear{Imbens and Rubin}{Imbens and Rubin}{2015}]%
        {imbens2015causal}
\bibfield{author}{\bibinfo{person}{Guido~W Imbens} {and}
  \bibinfo{person}{Donald~B Rubin}.} \bibinfo{year}{2015}\natexlab{}.
\newblock \bibinfo{booktitle}{\emph{Causal inference in statistics, social, and
  biomedical sciences}}.
\newblock \bibinfo{publisher}{Cambridge University Press}.
\newblock


\bibitem[\protect\citeauthoryear{Jacobs}{Jacobs}{2016}]%
        {jacobs2016death}
\bibfield{author}{\bibinfo{person}{Jane Jacobs}.}
  \bibinfo{year}{2016}\natexlab{}.
\newblock \bibinfo{booktitle}{\emph{The death and life of great American
  cities}}.
\newblock \bibinfo{publisher}{Vintage}.
\newblock


\bibitem[\protect\citeauthoryear{Kadar and Pletikosa}{Kadar and
  Pletikosa}{2018}]%
        {kadar2018mining}
\bibfield{author}{\bibinfo{person}{Cristina Kadar} {and} \bibinfo{person}{Irena
  Pletikosa}.} \bibinfo{year}{2018}\natexlab{}.
\newblock \showarticletitle{Mining large-scale human mobility data for
  long-term crime prediction}.
\newblock \bibinfo{journal}{\emph{EPJ Data Science}} \bibinfo{volume}{7},
  \bibinfo{number}{1} (\bibinfo{year}{2018}), \bibinfo{pages}{1--27}.
\newblock


\bibitem[\protect\citeauthoryear{Konda and Tsitsiklis}{Konda and
  Tsitsiklis}{2000}]%
        {konda2000actor}
\bibfield{author}{\bibinfo{person}{Vijay~R Konda} {and} \bibinfo{person}{John~N
  Tsitsiklis}.} \bibinfo{year}{2000}\natexlab{}.
\newblock \showarticletitle{Actor-critic algorithms}. In
  \bibinfo{booktitle}{\emph{Advances in neural information processing
  systems}}. \bibinfo{pages}{1008--1014}.
\newblock


\bibitem[\protect\citeauthoryear{Kriegel, Kr\"{o}ger, and Zimek}{Kriegel
  et~al\mbox{.}}{2009}]%
        {10.1145/1497577.1497578}
\bibfield{author}{\bibinfo{person}{Hans-Peter Kriegel}, \bibinfo{person}{Peer
  Kr\"{o}ger}, {and} \bibinfo{person}{Arthur Zimek}.}
  \bibinfo{year}{2009}\natexlab{}.
\newblock \showarticletitle{Clustering High-Dimensional Data: A Survey on
  Subspace Clustering, Pattern-Based Clustering, and Correlation Clustering}.
\newblock \bibinfo{journal}{\emph{ACM Trans. Knowl. Discov. Data}}
  \bibinfo{volume}{3}, \bibinfo{number}{1}, Article \bibinfo{articleno}{1}
  (\bibinfo{date}{mar} \bibinfo{year}{2009}), \bibinfo{numpages}{58}~pages.
\newblock
\showISSN{1556-4681}
\urldef\tempurl%
\url{https://doi.org/10.1145/1497577.1497578}
\showDOI{\tempurl}


\bibitem[\protect\citeauthoryear{Kuang, Cui, Li, Jiang, Wang, Wu, and
  Yang}{Kuang et~al\mbox{.}}{2019}]%
        {10.1145/3365677}
\bibfield{author}{\bibinfo{person}{Kun Kuang}, \bibinfo{person}{Peng Cui},
  \bibinfo{person}{Bo Li}, \bibinfo{person}{Meng Jiang},
  \bibinfo{person}{Yashen Wang}, \bibinfo{person}{Fei Wu}, {and}
  \bibinfo{person}{Shiqiang Yang}.} \bibinfo{year}{2019}\natexlab{}.
\newblock \showarticletitle{Treatment Effect Estimation via Differentiated
  Confounder Balancing and Regression}.
\newblock \bibinfo{journal}{\emph{ACM Trans. Knowl. Discov. Data}}
  \bibinfo{volume}{14}, \bibinfo{number}{1}, Article \bibinfo{articleno}{6}
  (\bibinfo{date}{dec} \bibinfo{year}{2019}), \bibinfo{numpages}{25}~pages.
\newblock
\showISSN{1556-4681}
\urldef\tempurl%
\url{https://doi.org/10.1145/3365677}
\showDOI{\tempurl}


\bibitem[\protect\citeauthoryear{Kuha}{Kuha}{2004}]%
        {kuha2004aic}
\bibfield{author}{\bibinfo{person}{Jouni Kuha}.}
  \bibinfo{year}{2004}\natexlab{}.
\newblock \showarticletitle{AIC and BIC: Comparisons of assumptions and
  performance}.
\newblock \bibinfo{journal}{\emph{Sociological methods \& research}}
  \bibinfo{volume}{33}, \bibinfo{number}{2} (\bibinfo{year}{2004}),
  \bibinfo{pages}{188--229}.
\newblock


\bibitem[\protect\citeauthoryear{Li, Wang, Wang, and Xu}{Li
  et~al\mbox{.}}{2022}]%
        {10.1145/3533725}
\bibfield{author}{\bibinfo{person}{Qian Li}, \bibinfo{person}{Xiangmeng Wang},
  \bibinfo{person}{Zhichao Wang}, {and} \bibinfo{person}{Guandong Xu}.}
  \bibinfo{year}{2022}\natexlab{}.
\newblock \showarticletitle{Be Causal: De-Biasing Social Network Confounding in
  Recommendation}.
\newblock \bibinfo{journal}{\emph{ACM Trans. Knowl. Discov. Data}}
  (\bibinfo{date}{apr} \bibinfo{year}{2022}).
\newblock
\showISSN{1556-4681}
\urldef\tempurl%
\url{https://doi.org/10.1145/3533725}
\showDOI{\tempurl}
\newblock
\shownote{Just Accepted.}


\bibitem[\protect\citeauthoryear{Liu, Liu, Lu, Teng, Zhu, and Xiong}{Liu
  et~al\mbox{.}}{2017}]%
        {liu2017point}
\bibfield{author}{\bibinfo{person}{Yanchi Liu}, \bibinfo{person}{Chuanren Liu},
  \bibinfo{person}{Xinjiang Lu}, \bibinfo{person}{Mingfei Teng},
  \bibinfo{person}{Hengshu Zhu}, {and} \bibinfo{person}{Hui Xiong}.}
  \bibinfo{year}{2017}\natexlab{}.
\newblock \showarticletitle{Point-of-interest demand modeling with human
  mobility patterns}. In \bibinfo{booktitle}{\emph{Proceedings of the 23rd ACM
  SIGKDD international conference on knowledge discovery and data mining}}.
  \bibinfo{pages}{947--955}.
\newblock


\bibitem[\protect\citeauthoryear{Lu, Zanutto, Hornik, and Rosenbaum}{Lu
  et~al\mbox{.}}{2001}]%
        {lu2001matching}
\bibfield{author}{\bibinfo{person}{Bo Lu}, \bibinfo{person}{Elaine Zanutto},
  \bibinfo{person}{Robert Hornik}, {and} \bibinfo{person}{Paul~R Rosenbaum}.}
  \bibinfo{year}{2001}\natexlab{}.
\newblock \showarticletitle{Matching with doses in an observational study of a
  media campaign against drug abuse}.
\newblock \bibinfo{journal}{\emph{J. Amer. Statist. Assoc.}}
  \bibinfo{volume}{96}, \bibinfo{number}{456} (\bibinfo{year}{2001}),
  \bibinfo{pages}{1245--1253}.
\newblock


\bibitem[\protect\citeauthoryear{Lu, Feng, Zhou, Li, and Cao}{Lu
  et~al\mbox{.}}{2018}]%
        {lu2018inferring}
\bibfield{author}{\bibinfo{person}{Zheng Lu}, \bibinfo{person}{Yunhe Feng},
  \bibinfo{person}{Wenjun Zhou}, \bibinfo{person}{Xiaolin Li}, {and}
  \bibinfo{person}{Qing Cao}.} \bibinfo{year}{2018}\natexlab{}.
\newblock \showarticletitle{Inferring correlation between user mobility and app
  usage in massive coarse-grained data traces}.
\newblock \bibinfo{journal}{\emph{Proceedings of the ACM on Interactive,
  Mobile, Wearable and Ubiquitous Technologies}} \bibinfo{volume}{1},
  \bibinfo{number}{4} (\bibinfo{year}{2018}), \bibinfo{pages}{1--21}.
\newblock


\bibitem[\protect\citeauthoryear{MacPherson, Gushulak, and
  Macdonald}{MacPherson et~al\mbox{.}}{2007}]%
        {macpherson2007health}
\bibfield{author}{\bibinfo{person}{Douglas~W MacPherson},
  \bibinfo{person}{Brian~D Gushulak}, {and} \bibinfo{person}{Liane Macdonald}.}
  \bibinfo{year}{2007}\natexlab{}.
\newblock \showarticletitle{Health and foreign policy: influences of migration
  and population mobility}.
\newblock \bibinfo{journal}{\emph{Bulletin of the World Health Organization}}
  \bibinfo{volume}{85} (\bibinfo{year}{2007}), \bibinfo{pages}{200--206}.
\newblock


\bibitem[\protect\citeauthoryear{Malinsky and Danks}{Malinsky and
  Danks}{2018}]%
        {malinsky2018causal}
\bibfield{author}{\bibinfo{person}{Daniel Malinsky} {and}
  \bibinfo{person}{David Danks}.} \bibinfo{year}{2018}\natexlab{}.
\newblock \showarticletitle{Causal discovery algorithms: A practical guide}.
\newblock \bibinfo{journal}{\emph{Philosophy Compass}} \bibinfo{volume}{13},
  \bibinfo{number}{1} (\bibinfo{year}{2018}), \bibinfo{pages}{e12470}.
\newblock


\bibitem[\protect\citeauthoryear{Mazyavkina, Sviridov, Ivanov, and
  Burnaev}{Mazyavkina et~al\mbox{.}}{2021}]%
        {mazyavkina2021reinforcement}
\bibfield{author}{\bibinfo{person}{Nina Mazyavkina}, \bibinfo{person}{Sergey
  Sviridov}, \bibinfo{person}{Sergei Ivanov}, {and} \bibinfo{person}{Evgeny
  Burnaev}.} \bibinfo{year}{2021}\natexlab{}.
\newblock \showarticletitle{Reinforcement learning for combinatorial
  optimization: A survey}.
\newblock \bibinfo{journal}{\emph{Computers \& Operations Research}}
  \bibinfo{volume}{134} (\bibinfo{year}{2021}), \bibinfo{pages}{105400}.
\newblock


\bibitem[\protect\citeauthoryear{McNamee}{McNamee}{2003}]%
        {mcnamee2003confounding}
\bibfield{author}{\bibinfo{person}{Roseanne McNamee}.}
  \bibinfo{year}{2003}\natexlab{}.
\newblock \showarticletitle{Confounding and confounders}.
\newblock \bibinfo{journal}{\emph{Occupational and environmental medicine}}
  \bibinfo{volume}{60}, \bibinfo{number}{3} (\bibinfo{year}{2003}),
  \bibinfo{pages}{227--234}.
\newblock


\bibitem[\protect\citeauthoryear{Metz}{Metz}{2000}]%
        {metz2000mobility}
\bibfield{author}{\bibinfo{person}{David~H Metz}.}
  \bibinfo{year}{2000}\natexlab{}.
\newblock \showarticletitle{Mobility of older people and their quality of
  life}.
\newblock \bibinfo{journal}{\emph{Transport policy}} \bibinfo{volume}{7},
  \bibinfo{number}{2} (\bibinfo{year}{2000}), \bibinfo{pages}{149--152}.
\newblock


\bibitem[\protect\citeauthoryear{M{\o}nsted, Mollgaard, and
  Mathiesen}{M{\o}nsted et~al\mbox{.}}{2018}]%
        {monsted2018phone}
\bibfield{author}{\bibinfo{person}{Bjarke M{\o}nsted}, \bibinfo{person}{Anders
  Mollgaard}, {and} \bibinfo{person}{Joachim Mathiesen}.}
  \bibinfo{year}{2018}\natexlab{}.
\newblock \showarticletitle{Phone-based metric as a predictor for basic
  personality traits}.
\newblock \bibinfo{journal}{\emph{Journal of Research in Personality}}
  \bibinfo{volume}{74} (\bibinfo{year}{2018}), \bibinfo{pages}{16--22}.
\newblock


\bibitem[\protect\citeauthoryear{Moro, Calacci, Dong, and Pentland}{Moro
  et~al\mbox{.}}{2021}]%
        {moro2021mobility}
\bibfield{author}{\bibinfo{person}{Esteban Moro}, \bibinfo{person}{Dan
  Calacci}, \bibinfo{person}{Xiaowen Dong}, {and} \bibinfo{person}{Alex
  Pentland}.} \bibinfo{year}{2021}\natexlab{}.
\newblock \showarticletitle{Mobility patterns are associated with experienced
  income segregation in large US cities}.
\newblock \bibinfo{journal}{\emph{Nature communications}} \bibinfo{volume}{12},
  \bibinfo{number}{1} (\bibinfo{year}{2021}), \bibinfo{pages}{1--10}.
\newblock


\bibitem[\protect\citeauthoryear{Myrovali, Karakasidis, Ayfantopoulou, and
  Morfoulaki}{Myrovali et~al\mbox{.}}{2022}]%
        {myrovali2022spatio}
\bibfield{author}{\bibinfo{person}{Glykeria Myrovali},
  \bibinfo{person}{Theodoros Karakasidis}, \bibinfo{person}{Georgia
  Ayfantopoulou}, {and} \bibinfo{person}{Maria Morfoulaki}.}
  \bibinfo{year}{2022}\natexlab{}.
\newblock \showarticletitle{Spatio-Temporal Causal Relations at Urban Road
  Networks; Granger Causality Based Networks as an Insight to Urban Traffic
  Dynamics}. In \bibinfo{booktitle}{\emph{Proceedings of Sixth International
  Congress on Information and Communication Technology}}. Springer,
  \bibinfo{pages}{791--804}.
\newblock


\bibitem[\protect\citeauthoryear{Nevelsteen, Steenberghen, Van~Rompaey, and
  Uyttersprot}{Nevelsteen et~al\mbox{.}}{2012}]%
        {nevelsteen2012controlling}
\bibfield{author}{\bibinfo{person}{Kristof Nevelsteen},
  \bibinfo{person}{Th{\'e}r{\`e}se Steenberghen}, \bibinfo{person}{Anton
  Van~Rompaey}, {and} \bibinfo{person}{Liesbeth Uyttersprot}.}
  \bibinfo{year}{2012}\natexlab{}.
\newblock \showarticletitle{Controlling factors of the parental safety
  perception on children's travel mode choice}.
\newblock \bibinfo{journal}{\emph{Accident Analysis \& Prevention}}
  \bibinfo{volume}{45} (\bibinfo{year}{2012}), \bibinfo{pages}{39--49}.
\newblock


\bibitem[\protect\citeauthoryear{{OpenStreetMap contributors}}{{OpenStreetMap
  contributors}}{2017}]%
        {OpenStreetMap}
\bibfield{author}{\bibinfo{person}{{OpenStreetMap contributors}}.}
  \bibinfo{year}{2017}\natexlab{}.
\newblock \bibinfo{title}{{Planet dump retrieved from https://planet.osm.org
  }}.
\newblock \bibinfo{howpublished}{\url{ https://www.openstreetmap.org }}.
\newblock


\bibitem[\protect\citeauthoryear{Pearl}{Pearl}{2009}]%
        {pearl2009causal}
\bibfield{author}{\bibinfo{person}{Judea Pearl}.}
  \bibinfo{year}{2009}\natexlab{}.
\newblock \showarticletitle{Causal inference in statistics: An overview}.
\newblock \bibinfo{journal}{\emph{Statistics surveys}}  \bibinfo{volume}{3}
  (\bibinfo{year}{2009}), \bibinfo{pages}{96--146}.
\newblock


\bibitem[\protect\citeauthoryear{Pearl}{Pearl}{2010}]%
        {pearl2010causal}
\bibfield{author}{\bibinfo{person}{Judea Pearl}.}
  \bibinfo{year}{2010}\natexlab{}.
\newblock \showarticletitle{Causal inference}.
\newblock \bibinfo{journal}{\emph{Causality: objectives and assessment}}
  (\bibinfo{year}{2010}), \bibinfo{pages}{39--58}.
\newblock


\bibitem[\protect\citeauthoryear{Pearl et~al\mbox{.}}{Pearl
  et~al\mbox{.}}{2000}]%
        {pearl2000models}
\bibfield{author}{\bibinfo{person}{Judea Pearl} {et~al\mbox{.}}}
  \bibinfo{year}{2000}\natexlab{}.
\newblock \showarticletitle{Models, reasoning and inference}.
\newblock \bibinfo{journal}{\emph{Cambridge, UK: CambridgeUniversityPress}}
  \bibinfo{volume}{19}, \bibinfo{number}{2} (\bibinfo{year}{2000}).
\newblock


\bibitem[\protect\citeauthoryear{Philips}{Philips}{2000}]%
        {philips2000shopping}
\bibfield{author}{\bibinfo{person}{Deborah Philips}.}
  \bibinfo{year}{2000}\natexlab{}.
\newblock \showarticletitle{Shopping for men: the single woman narrative}.
\newblock \bibinfo{journal}{\emph{Women: a cultural review}}
  \bibinfo{volume}{11}, \bibinfo{number}{3} (\bibinfo{year}{2000}),
  \bibinfo{pages}{238--251}.
\newblock


\bibitem[\protect\citeauthoryear{Portugali}{Portugali}{2000}]%
        {portugali2000self}
\bibfield{author}{\bibinfo{person}{Juval Portugali}.}
  \bibinfo{year}{2000}\natexlab{}.
\newblock \bibinfo{booktitle}{\emph{Self-organization and the city}}.
\newblock \bibinfo{publisher}{Springer Science \& Business Media}.
\newblock


\bibitem[\protect\citeauthoryear{Rauws, Cozzolino, and Moroni}{Rauws
  et~al\mbox{.}}{2020}]%
        {rauws2020framework}
\bibfield{author}{\bibinfo{person}{Ward Rauws}, \bibinfo{person}{Stefano
  Cozzolino}, {and} \bibinfo{person}{Stefano Moroni}.}
  \bibinfo{year}{2020}\natexlab{}.
\newblock \bibinfo{title}{Framework rules for self-organizing cities:
  Introduction}.
\newblock
\newblock


\bibitem[\protect\citeauthoryear{Rosenbaum and Rubin}{Rosenbaum and
  Rubin}{1983}]%
        {rosenbaum1983central}
\bibfield{author}{\bibinfo{person}{Paul~R Rosenbaum} {and}
  \bibinfo{person}{Donald~B Rubin}.} \bibinfo{year}{1983}\natexlab{}.
\newblock \showarticletitle{The central role of the propensity score in
  observational studies for causal effects}.
\newblock \bibinfo{journal}{\emph{Biometrika}} \bibinfo{volume}{70},
  \bibinfo{number}{1} (\bibinfo{year}{1983}), \bibinfo{pages}{41--55}.
\newblock


\bibitem[\protect\citeauthoryear{Ruan, Bao, Liang, Li, He, Meng, Li, Wu, and
  Zheng}{Ruan et~al\mbox{.}}{2020}]%
        {ruan2020dynamic}
\bibfield{author}{\bibinfo{person}{Sijie Ruan}, \bibinfo{person}{Jie Bao},
  \bibinfo{person}{Yuxuan Liang}, \bibinfo{person}{Ruiyuan Li},
  \bibinfo{person}{Tianfu He}, \bibinfo{person}{Chuishi Meng},
  \bibinfo{person}{Yanhua Li}, \bibinfo{person}{Yingcai Wu}, {and}
  \bibinfo{person}{Yu Zheng}.} \bibinfo{year}{2020}\natexlab{}.
\newblock \showarticletitle{Dynamic Public Resource Allocation Based on Human
  Mobility Prediction}.
\newblock \bibinfo{journal}{\emph{Proceedings of the ACM on Interactive,
  Mobile, Wearable and Ubiquitous Technologies}} \bibinfo{volume}{4},
  \bibinfo{number}{1} (\bibinfo{year}{2020}), \bibinfo{pages}{1--22}.
\newblock


\bibitem[\protect\citeauthoryear{Rubin}{Rubin}{1974}]%
        {rubin1974estimating}
\bibfield{author}{\bibinfo{person}{Donald~B Rubin}.}
  \bibinfo{year}{1974}\natexlab{}.
\newblock \showarticletitle{Estimating causal effects of treatments in
  randomized and nonrandomized studies.}
\newblock \bibinfo{journal}{\emph{Journal of educational Psychology}}
  \bibinfo{volume}{66}, \bibinfo{number}{5} (\bibinfo{year}{1974}),
  \bibinfo{pages}{688}.
\newblock


\bibitem[\protect\citeauthoryear{Sadri, Shahriari~Ahmadi, and Tajalli}{Sadri
  et~al\mbox{.}}{2021}]%
        {sadri2021causal}
\bibfield{author}{\bibinfo{person}{Leila Sadri}, \bibinfo{person}{Mansoureh
  Shahriari~Ahmadi}, {and} \bibinfo{person}{Parisa Tajalli}.}
  \bibinfo{year}{2021}\natexlab{}.
\newblock \showarticletitle{Causal Relationship between Peer-Matching and Body
  Management with Mediating Role of Mobile-Based Social Media Addiction in
  Adolescents with Social Anxiety Leila Sadri Mansooreh Shahriari Ahmadi*
  Parisa Tajali}.
\newblock \bibinfo{journal}{\emph{Quarterly Social Psychology Research}}
  \bibinfo{volume}{11}, \bibinfo{number}{41} (\bibinfo{year}{2021}),
  \bibinfo{pages}{121--136}.
\newblock


\bibitem[\protect\citeauthoryear{Shi, Shen, Kou, Nie, and Yu}{Shi
  et~al\mbox{.}}{2021}]%
        {shi2021attentional}
\bibfield{author}{\bibinfo{person}{Meihui Shi}, \bibinfo{person}{Derong Shen},
  \bibinfo{person}{Yue Kou}, \bibinfo{person}{Tiezheng Nie}, {and}
  \bibinfo{person}{Ge Yu}.} \bibinfo{year}{2021}\natexlab{}.
\newblock \showarticletitle{Attentional Memory Network with Correlation-based
  Embedding for time-aware POI recommendation}.
\newblock \bibinfo{journal}{\emph{Knowledge-Based Systems}}
  \bibinfo{volume}{214} (\bibinfo{year}{2021}), \bibinfo{pages}{106747}.
\newblock


\bibitem[\protect\citeauthoryear{Shimosaka, Maeda, Tsukiji, and
  Tsubouchi}{Shimosaka et~al\mbox{.}}{2015}]%
        {shimosaka2015forecasting}
\bibfield{author}{\bibinfo{person}{Masamichi Shimosaka},
  \bibinfo{person}{Keisuke Maeda}, \bibinfo{person}{Takeshi Tsukiji}, {and}
  \bibinfo{person}{Kota Tsubouchi}.} \bibinfo{year}{2015}\natexlab{}.
\newblock \showarticletitle{Forecasting urban dynamics with mobility logs by
  bilinear Poisson regression}. In \bibinfo{booktitle}{\emph{Proceedings of the
  2015 ACM international joint conference on pervasive and ubiquitous
  computing}}. \bibinfo{pages}{535--546}.
\newblock


\bibitem[\protect\citeauthoryear{Smith, Kajumba, Bobholz, Smith, Kaddumukasa,
  Kakooza-Mwesige, Chakraborty, Sinha, Kaddumukasa, Gualtieri,
  et~al\mbox{.}}{Smith et~al\mbox{.}}{2021}]%
        {smith2021pluralistic}
\bibfield{author}{\bibinfo{person}{Caleigh~E Smith}, \bibinfo{person}{Mayanja
  Kajumba}, \bibinfo{person}{Samuel Bobholz}, \bibinfo{person}{Patrick~J
  Smith}, \bibinfo{person}{Mark Kaddumukasa}, \bibinfo{person}{Angelina
  Kakooza-Mwesige}, \bibinfo{person}{Payal Chakraborty},
  \bibinfo{person}{Drishti~D Sinha}, \bibinfo{person}{Martin~N Kaddumukasa},
  \bibinfo{person}{Alex Gualtieri}, {et~al\mbox{.}}}
  \bibinfo{year}{2021}\natexlab{}.
\newblock \showarticletitle{Pluralistic and singular causal attributions for
  epilepsy in Uganda}.
\newblock \bibinfo{journal}{\emph{Epilepsy \& Behavior}}  \bibinfo{volume}{114}
  (\bibinfo{year}{2021}), \bibinfo{pages}{107334}.
\newblock


\bibitem[\protect\citeauthoryear{Spirtes and Zhang}{Spirtes and Zhang}{2016}]%
        {spirtes2016causal}
\bibfield{author}{\bibinfo{person}{Peter Spirtes} {and} \bibinfo{person}{Kun
  Zhang}.} \bibinfo{year}{2016}\natexlab{}.
\newblock \showarticletitle{Causal discovery and inference: concepts and recent
  methodological advances}. In \bibinfo{booktitle}{\emph{Applied informatics}},
  Vol.~\bibinfo{volume}{3}. SpringerOpen, \bibinfo{pages}{1--28}.
\newblock


\bibitem[\protect\citeauthoryear{Tibshirani}{Tibshirani}{1996}]%
        {tibshirani1996regression}
\bibfield{author}{\bibinfo{person}{Robert Tibshirani}.}
  \bibinfo{year}{1996}\natexlab{}.
\newblock \showarticletitle{Regression shrinkage and selection via the lasso}.
\newblock \bibinfo{journal}{\emph{Journal of the Royal Statistical Society:
  Series B (Methodological)}} \bibinfo{volume}{58}, \bibinfo{number}{1}
  (\bibinfo{year}{1996}), \bibinfo{pages}{267--288}.
\newblock


\bibitem[\protect\citeauthoryear{UN}{UN}{2015}]%
        {un2018world}
\bibfield{author}{\bibinfo{person}{DESA UN}.} \bibinfo{year}{2015}\natexlab{}.
\newblock \showarticletitle{World Urbanization Prospects: The 2018 Revision.}
\newblock \bibinfo{journal}{\emph{United Nations Department of Economics and
  Social Affairs, Population Division: New York, NY, USA}}
  \bibinfo{volume}{41} (\bibinfo{year}{2015}).
\newblock


\bibitem[\protect\citeauthoryear{{United States Census Bureau
  contributors}}{{United States Census Bureau contributors}}{2020}]%
        {USCB}
\bibfield{author}{\bibinfo{person}{{United States Census Bureau
  contributors}}.} \bibinfo{year}{2020}\natexlab{}.
\newblock \bibinfo{title}{{Planet dump retrieved from https://www.census.gov
  }}.
\newblock \bibinfo{howpublished}{\url{https://www.census.gov}}.
\newblock


\bibitem[\protect\citeauthoryear{VanderWeele and Shpitser}{VanderWeele and
  Shpitser}{2013}]%
        {vanderweele2013definition}
\bibfield{author}{\bibinfo{person}{Tyler~J VanderWeele} {and}
  \bibinfo{person}{Ilya Shpitser}.} \bibinfo{year}{2013}\natexlab{}.
\newblock \showarticletitle{On the definition of a confounder}.
\newblock \bibinfo{journal}{\emph{Annals of statistics}} \bibinfo{volume}{41},
  \bibinfo{number}{1} (\bibinfo{year}{2013}), \bibinfo{pages}{196}.
\newblock


\bibitem[\protect\citeauthoryear{Wang and Li}{Wang and Li}{2017}]%
        {wang2017region}
\bibfield{author}{\bibinfo{person}{Hongjian Wang} {and}
  \bibinfo{person}{Zhenhui Li}.} \bibinfo{year}{2017}\natexlab{}.
\newblock \showarticletitle{Region representation learning via mobility flow}.
  In \bibinfo{booktitle}{\emph{Proceedings of the 2017 ACM on Conference on
  Information and Knowledge Management}}. \bibinfo{pages}{237--246}.
\newblock


\bibitem[\protect\citeauthoryear{Wang, Du, Zhu, Ke, Chen, Hao, and Wang}{Wang
  et~al\mbox{.}}{2021}]%
        {wang2021ordering}
\bibfield{author}{\bibinfo{person}{Xiaoqiang Wang}, \bibinfo{person}{Yali Du},
  \bibinfo{person}{Shengyu Zhu}, \bibinfo{person}{Liangjun Ke},
  \bibinfo{person}{Zhitang Chen}, \bibinfo{person}{Jianye Hao}, {and}
  \bibinfo{person}{Jun Wang}.} \bibinfo{year}{2021}\natexlab{}.
\newblock \showarticletitle{Ordering-Based Causal Discovery with Reinforcement
  Learning}.
\newblock \bibinfo{journal}{\emph{arXiv preprint arXiv:2105.06631}}
  (\bibinfo{year}{2021}).
\newblock


\bibitem[\protect\citeauthoryear{Wang, Song, Du, and Lu}{Wang
  et~al\mbox{.}}{2019}]%
        {10.1145/3356584}
\bibfield{author}{\bibinfo{person}{Yun Wang}, \bibinfo{person}{Guojie Song},
  \bibinfo{person}{Lun Du}, {and} \bibinfo{person}{Zhicong Lu}.}
  \bibinfo{year}{2019}\natexlab{}.
\newblock \showarticletitle{Real-Time Estimation of the Urban Air Quality with
  Mobile Sensor System}.
\newblock \bibinfo{journal}{\emph{ACM Trans. Knowl. Discov. Data}}
  \bibinfo{volume}{13}, \bibinfo{number}{5}, Article \bibinfo{articleno}{49}
  (\bibinfo{date}{sep} \bibinfo{year}{2019}), \bibinfo{numpages}{9}~pages.
\newblock
\showISSN{1556-4681}
\urldef\tempurl%
\url{https://doi.org/10.1145/3356584}
\showDOI{\tempurl}


\bibitem[\protect\citeauthoryear{Weidlich}{Weidlich}{1999}]%
        {weidlich1999fast}
\bibfield{author}{\bibinfo{person}{Wolfgang Weidlich}.}
  \bibinfo{year}{1999}\natexlab{}.
\newblock \showarticletitle{From fast to slow processes in the evolution of
  urban and regional settlement structures}.
\newblock \bibinfo{journal}{\emph{Discrete Dynamics in Nature and Society}}
  \bibinfo{volume}{3}, \bibinfo{number}{2-3} (\bibinfo{year}{1999}),
  \bibinfo{pages}{137--147}.
\newblock


\bibitem[\protect\citeauthoryear{Xu, Lin, Xia, Guo, and Li}{Xu
  et~al\mbox{.}}{2020a}]%
        {xu2020sume}
\bibfield{author}{\bibinfo{person}{Fengli Xu}, \bibinfo{person}{Zongyu Lin},
  \bibinfo{person}{Tong Xia}, \bibinfo{person}{Diansheng Guo}, {and}
  \bibinfo{person}{Yong Li}.} \bibinfo{year}{2020}\natexlab{a}.
\newblock \showarticletitle{Sume: Semantic-enhanced urban mobility network
  embedding for user demographic inference}.
\newblock \bibinfo{journal}{\emph{Proceedings of the ACM on Interactive,
  Mobile, Wearable and Ubiquitous Technologies}} \bibinfo{volume}{4},
  \bibinfo{number}{3} (\bibinfo{year}{2020}), \bibinfo{pages}{1--25}.
\newblock


\bibitem[\protect\citeauthoryear{Xu, Ge, Li, Fu, Chen, and Zhang}{Xu
  et~al\mbox{.}}{2021a}]%
        {xu2021causal}
\bibfield{author}{\bibinfo{person}{Shuyuan Xu}, \bibinfo{person}{Yingqiang Ge},
  \bibinfo{person}{Yunqi Li}, \bibinfo{person}{Zuohui Fu}, \bibinfo{person}{Xu
  Chen}, {and} \bibinfo{person}{Yongfeng Zhang}.}
  \bibinfo{year}{2021}\natexlab{a}.
\newblock \showarticletitle{Causal Collaborative Filtering}.
\newblock \bibinfo{journal}{\emph{arXiv preprint arXiv:2102.01868}}
  (\bibinfo{year}{2021}).
\newblock


\bibitem[\protect\citeauthoryear{Xu, Li, Liu, Fu, Ge, Chen, and Zhang}{Xu
  et~al\mbox{.}}{2021b}]%
        {xu2021learning}
\bibfield{author}{\bibinfo{person}{Shuyuan Xu}, \bibinfo{person}{Yunqi Li},
  \bibinfo{person}{Shuchang Liu}, \bibinfo{person}{Zuohui Fu},
  \bibinfo{person}{Yingqiang Ge}, \bibinfo{person}{Xu Chen}, {and}
  \bibinfo{person}{Yongfeng Zhang}.} \bibinfo{year}{2021}\natexlab{b}.
\newblock \showarticletitle{Learning Causal Explanations for Recommendation}.
  In \bibinfo{booktitle}{\emph{The 1st International Workshop on Causality in
  Search and Recommendation}}.
\newblock


\bibitem[\protect\citeauthoryear{Xu, Olmos, Abbar, and Gonz{\'a}lez}{Xu
  et~al\mbox{.}}{2020b}]%
        {xu2020deconstructing}
\bibfield{author}{\bibinfo{person}{Yanyan Xu}, \bibinfo{person}{Luis~E Olmos},
  \bibinfo{person}{Sofiane Abbar}, {and} \bibinfo{person}{Marta~C
  Gonz{\'a}lez}.} \bibinfo{year}{2020}\natexlab{b}.
\newblock \showarticletitle{Deconstructing laws of accessibility and facility
  distribution in cities}.
\newblock \bibinfo{journal}{\emph{Science advances}} \bibinfo{volume}{6},
  \bibinfo{number}{37} (\bibinfo{year}{2020}), \bibinfo{pages}{eabb4112}.
\newblock


\bibitem[\protect\citeauthoryear{Xu, Zhu, Shen, and Yu}{Xu
  et~al\mbox{.}}{2019}]%
        {10.1145/3340847}
\bibfield{author}{\bibinfo{person}{Yanan Xu}, \bibinfo{person}{Yanmin Zhu},
  \bibinfo{person}{Yanyan Shen}, {and} \bibinfo{person}{Jiadi Yu}.}
  \bibinfo{year}{2019}\natexlab{}.
\newblock \showarticletitle{Fine-Grained Air Quality Inference with Remote
  Sensing Data and Ubiquitous Urban Data}.
\newblock \bibinfo{journal}{\emph{ACM Trans. Knowl. Discov. Data}}
  \bibinfo{volume}{13}, \bibinfo{number}{5}, Article \bibinfo{articleno}{46}
  (\bibinfo{date}{sep} \bibinfo{year}{2019}), \bibinfo{numpages}{27}~pages.
\newblock
\showISSN{1556-4681}
\urldef\tempurl%
\url{https://doi.org/10.1145/3340847}
\showDOI{\tempurl}


\bibitem[\protect\citeauthoryear{Yao, Chu, Li, Li, Gao, and Zhang}{Yao
  et~al\mbox{.}}{2021}]%
        {10.1145/3444944}
\bibfield{author}{\bibinfo{person}{Liuyi Yao}, \bibinfo{person}{Zhixuan Chu},
  \bibinfo{person}{Sheng Li}, \bibinfo{person}{Yaliang Li},
  \bibinfo{person}{Jing Gao}, {and} \bibinfo{person}{Aidong Zhang}.}
  \bibinfo{year}{2021}\natexlab{}.
\newblock \showarticletitle{A Survey on Causal Inference}.
\newblock \bibinfo{journal}{\emph{ACM Trans. Knowl. Discov. Data}}
  \bibinfo{volume}{15}, \bibinfo{number}{5}, Article \bibinfo{articleno}{74}
  (\bibinfo{date}{may} \bibinfo{year}{2021}), \bibinfo{numpages}{46}~pages.
\newblock
\showISSN{1556-4681}
\urldef\tempurl%
\url{https://doi.org/10.1145/3444944}
\showDOI{\tempurl}


\bibitem[\protect\citeauthoryear{Yao}{Yao}{2018}]%
        {yao2018exploiting}
\bibfield{author}{\bibinfo{person}{Zijun Yao}.}
  \bibinfo{year}{2018}\natexlab{}.
\newblock \showarticletitle{Exploiting human mobility patterns for
  point-of-interest recommendation}. In \bibinfo{booktitle}{\emph{Proceedings
  of the Eleventh ACM International Conference on Web Search and Data Mining}}.
  \bibinfo{pages}{757--758}.
\newblock


\bibitem[\protect\citeauthoryear{Yue, Zhuang, Yeh, Xie, Ma, and Li}{Yue
  et~al\mbox{.}}{2017}]%
        {yue2017measurements}
\bibfield{author}{\bibinfo{person}{Yang Yue}, \bibinfo{person}{Yan Zhuang},
  \bibinfo{person}{Anthony~GO Yeh}, \bibinfo{person}{Jin-Yun Xie},
  \bibinfo{person}{Cheng-Lin Ma}, {and} \bibinfo{person}{Qing-Quan Li}.}
  \bibinfo{year}{2017}\natexlab{}.
\newblock \showarticletitle{Measurements of POI-based mixed use and their
  relationships with neighbourhood vibrancy}.
\newblock \bibinfo{journal}{\emph{International Journal of Geographical
  Information Science}} \bibinfo{volume}{31}, \bibinfo{number}{4}
  (\bibinfo{year}{2017}), \bibinfo{pages}{658--675}.
\newblock


\bibitem[\protect\citeauthoryear{Zeng, Fu, Arisona, Schubiger, Burkhard, and
  Ma}{Zeng et~al\mbox{.}}{2017}]%
        {zeng2017visualizing}
\bibfield{author}{\bibinfo{person}{Wei Zeng}, \bibinfo{person}{Chi-Wing Fu},
  \bibinfo{person}{Stefan~M{\"u}ller Arisona}, \bibinfo{person}{Simon
  Schubiger}, \bibinfo{person}{Remo Burkhard}, {and} \bibinfo{person}{Kwan-Liu
  Ma}.} \bibinfo{year}{2017}\natexlab{}.
\newblock \showarticletitle{Visualizing the relationship between human mobility
  and points of interest}.
\newblock \bibinfo{journal}{\emph{IEEE Transactions on Intelligent
  Transportation Systems}} \bibinfo{volume}{18}, \bibinfo{number}{8}
  (\bibinfo{year}{2017}), \bibinfo{pages}{2271--2284}.
\newblock


\bibitem[\protect\citeauthoryear{Zhang, Wang, Xie, Ge, and Liu}{Zhang
  et~al\mbox{.}}{2020}]%
        {zhang2020real}
\bibfield{author}{\bibinfo{person}{Wen Zhang}, \bibinfo{person}{Yang Wang},
  \bibinfo{person}{Xike Xie}, \bibinfo{person}{Chuancai Ge}, {and}
  \bibinfo{person}{Hengchang Liu}.} \bibinfo{year}{2020}\natexlab{}.
\newblock \showarticletitle{Real-time Travel Time Estimation with Sparse
  Reliable Surveillance Information}.
\newblock \bibinfo{journal}{\emph{Proceedings of the ACM on Interactive,
  Mobile, Wearable and Ubiquitous Technologies}} \bibinfo{volume}{4},
  \bibinfo{number}{1} (\bibinfo{year}{2020}), \bibinfo{pages}{1--23}.
\newblock


\bibitem[\protect\citeauthoryear{Zhang, Xu, Xia, and Li}{Zhang
  et~al\mbox{.}}{2021}]%
        {zhang2021quantifying}
\bibfield{author}{\bibinfo{person}{Yunke Zhang}, \bibinfo{person}{Fengli Xu},
  \bibinfo{person}{Tong Xia}, {and} \bibinfo{person}{Yong Li}.}
  \bibinfo{year}{2021}\natexlab{}.
\newblock \showarticletitle{Quantifying the Causal Effect of Individual
  Mobility on Health Status in Urban Space}.
\newblock \bibinfo{journal}{\emph{Proceedings of the ACM on Interactive,
  Mobile, Wearable and Ubiquitous Technologies}} \bibinfo{volume}{5},
  \bibinfo{number}{4} (\bibinfo{year}{2021}), \bibinfo{pages}{1--30}.
\newblock


\bibitem[\protect\citeauthoryear{Zhao, Wang, Li, Zhang, and Yang}{Zhao
  et~al\mbox{.}}{2019}]%
        {zhao2019celltrans}
\bibfield{author}{\bibinfo{person}{Yi Zhao}, \bibinfo{person}{Xu Wang},
  \bibinfo{person}{Jianbo Li}, \bibinfo{person}{Desheng Zhang}, {and}
  \bibinfo{person}{Zheng Yang}.} \bibinfo{year}{2019}\natexlab{}.
\newblock \showarticletitle{CellTrans: Private Car or Public Transportation?
  Infer Users' Main Transportation Modes at Urban Scale with Cellular Data}.
\newblock \bibinfo{journal}{\emph{Proceedings of the ACM on Interactive,
  Mobile, Wearable and Ubiquitous Technologies}} \bibinfo{volume}{3},
  \bibinfo{number}{3} (\bibinfo{year}{2019}), \bibinfo{pages}{1--26}.
\newblock


\bibitem[\protect\citeauthoryear{Zheleva and Arbour}{Zheleva and
  Arbour}{2021}]%
        {zheleva2021causal}
\bibfield{author}{\bibinfo{person}{Elena Zheleva} {and} \bibinfo{person}{David
  Arbour}.} \bibinfo{year}{2021}\natexlab{}.
\newblock \showarticletitle{Causal Inference from Network Data}. In
  \bibinfo{booktitle}{\emph{Proceedings of the 27th ACM SIGKDD Conference on
  Knowledge Discovery \& Data Mining}}. \bibinfo{pages}{4096--4097}.
\newblock


\bibitem[\protect\citeauthoryear{Zheng, Aragam, Ravikumar, and Xing}{Zheng
  et~al\mbox{.}}{2018}]%
        {zheng2018dags}
\bibfield{author}{\bibinfo{person}{Xun Zheng}, \bibinfo{person}{Bryon Aragam},
  \bibinfo{person}{Pradeep Ravikumar}, {and} \bibinfo{person}{Eric~P Xing}.}
  \bibinfo{year}{2018}\natexlab{}.
\newblock \showarticletitle{Dags with no tears: Continuous optimization for
  structure learning}.
\newblock \bibinfo{journal}{\emph{arXiv preprint arXiv:1803.01422}}
  (\bibinfo{year}{2018}).
\newblock


\bibitem[\protect\citeauthoryear{Zheng, Capra, Wolfson, and Yang}{Zheng
  et~al\mbox{.}}{2014}]%
        {zheng2014urban}
\bibfield{author}{\bibinfo{person}{Yu Zheng}, \bibinfo{person}{Licia Capra},
  \bibinfo{person}{Ouri Wolfson}, {and} \bibinfo{person}{Hai Yang}.}
  \bibinfo{year}{2014}\natexlab{}.
\newblock \showarticletitle{Urban computing: concepts, methodologies, and
  applications}.
\newblock \bibinfo{journal}{\emph{ACM Transactions on Intelligent Systems and
  Technology (TIST)}} \bibinfo{volume}{5}, \bibinfo{number}{3}
  (\bibinfo{year}{2014}), \bibinfo{pages}{1--55}.
\newblock


\bibitem[\protect\citeauthoryear{Zheng, Gao, Li, He, Li, and Jin}{Zheng
  et~al\mbox{.}}{2021}]%
        {zheng2021disentangling}
\bibfield{author}{\bibinfo{person}{Yu Zheng}, \bibinfo{person}{Chen Gao},
  \bibinfo{person}{Xiang Li}, \bibinfo{person}{Xiangnan He},
  \bibinfo{person}{Yong Li}, {and} \bibinfo{person}{Depeng Jin}.}
  \bibinfo{year}{2021}\natexlab{}.
\newblock \showarticletitle{Disentangling User Interest and Conformity for
  Recommendation with Causal Embedding}. In
  \bibinfo{booktitle}{\emph{Proceedings of the Web Conference 2021}}.
  \bibinfo{pages}{2980--2991}.
\newblock


\bibitem[\protect\citeauthoryear{Zhong, Souza, and Mueen}{Zhong
  et~al\mbox{.}}{2022}]%
        {10.1145/3502738}
\bibfield{author}{\bibinfo{person}{Sheng Zhong}, \bibinfo{person}{Vinicius
  M.~A. Souza}, {and} \bibinfo{person}{Abdullah Mueen}.}
  \bibinfo{year}{2022}\natexlab{}.
\newblock \showarticletitle{Combining Filtering and Cross-Correlation
  Efficiently for Streaming Time Series}.
\newblock \bibinfo{journal}{\emph{ACM Trans. Knowl. Discov. Data}}
  \bibinfo{volume}{16}, \bibinfo{number}{5}, Article \bibinfo{articleno}{100}
  (\bibinfo{date}{may} \bibinfo{year}{2022}), \bibinfo{numpages}{24}~pages.
\newblock
\showISSN{1556-4681}
\urldef\tempurl%
\url{https://doi.org/10.1145/3502738}
\showDOI{\tempurl}


\bibitem[\protect\citeauthoryear{Zhu, Ng, and Chen}{Zhu et~al\mbox{.}}{2019}]%
        {zhu2019causal}
\bibfield{author}{\bibinfo{person}{Shengyu Zhu}, \bibinfo{person}{Ignavier Ng},
  {and} \bibinfo{person}{Zhitang Chen}.} \bibinfo{year}{2019}\natexlab{}.
\newblock \showarticletitle{Causal Discovery with Reinforcement Learning}. In
  \bibinfo{booktitle}{\emph{International Conference on Learning
  Representations}}.
\newblock


\end{thebibliography}


\end{document}